\documentclass[11pt]{article}

\usepackage[preprint]{acl}

\usepackage{times}
\usepackage{latexsym}
\usepackage{booktabs} 
\usepackage{multirow} 
\newsavebox{\promptbox} 
\newsavebox{\bprimebox} 
\definecolor{promptbg}{RGB}{223,231,243} 

\usepackage[T1]{fontenc}
\usepackage[utf8]{inputenc}

\usepackage{microtype}
\usepackage{inconsolata}
\usepackage{graphicx}
\usepackage{amsmath,amssymb,amsfonts}
\usepackage{algorithm} 
\usepackage{algpseudocode}
\usepackage{xcolor}

\title{How Far Can Root Cause Analysis Go on Real-World Telemetry Data?}

\author{Athira Gopal \\
 QPIAI India \\
 \texttt{athira.g@qpiai.tech} \\\And
 Ashwanth Krishnan \\
 QPIAI India \\
 \texttt{ashwanth.krishnan@qpiai.tech} \\}

\begin{document}
\maketitle

\begin{abstract}

Root cause analysis (RCA) of microservice failures requires reasoning over
telemetry far larger than a model's context, so a language agent reasons over
extracted evidence. We start from an error analysis of such an agent on
OpenRCA. A reverse reasoning agent reconstructs an evidence chain to each
labelled cause and compares it against the agent's own trace, locating failures
where a grounded chain existed but was not followed; its labels agree with two
site reliability engineers. Being label-conditioned, it audits evidence rather
than testing forward identifiability, leaving open what the agent lacked:
procedure or telemetry-specific knowledge. Existing systems cannot separate
them, extracting evidence inside the reasoning loop. We compute the evidence
once, offline, and supply it identically to every arm. Rules mined from the
audit reports of one Market cloudbed raise Full Score on a held-out cloudbed
from 14.10 to 35.90, surviving correction for multiple testing, against 24.36
for manual rules and 19.23 for a label-and-schema control. The rules add 17
solved cases on top of the five-step procedure but 11 without it, the repairs
overlapping. Removing the procedure raises Full Score in all three conditions
tested, though no contrast survives correction. Deleting only its
earliest-onset criterion yields a net gain of 8 cases against 11 for the whole
procedure, which we report descriptively. Mining is label-supervised, and all results are specific
to OpenRCA.

\end{abstract}

\section{Introduction}

Root cause analysis (RCA) on real-world telemetry data is challenging because of data volume,
heterogeneous formats and the absence of documented domain knowledge. Telemetry is spread across
many CSV files holding metrics, logs and traces, so the problem is inherently multimodal. The
OpenRCA benchmark \citep{xu2025openrca} exemplifies this and yields consistently low accuracy
across existing methods. Its task is to identify one or two root causes inside a 30-minute failure
window, each as a component, a fault reason and an onset timestamp. Extracting that half-hour from every telemetry file
yields roughly 2\,GB in our extraction, far beyond the context of current models, so a
system cannot be prompted with the window directly: it must first extract candidate evidence from the
window and then reason over what the extraction returns. How that
extraction is done, and what the model is told about how to read it, are
therefore design choices rather than details.

OpenRCA is organised into three datasets, Market, Telecom and Bank, which
differ in difficulty and in root-cause characteristics. Market alone is
split into two cloudbed deployments, CB1 and CB2, which share a topology
and fault vocabulary but differ in component instances, and that split is
what makes a held-out transfer experiment possible.

Existing systems extract their own evidence as they reason, and that is
where they are weak. The OpenRCA agent writes and executes code
over the raw telemetry, analysing metrics first and verifying with logs and traces; in the runs we observed it often
fails to examine all available files and tends to select the first detected
anomaly. Because
it computes anomalies inside the same reasoning loop rather than from a consistent precomputed set,
its detection is ad hoc. The remaining LLM-based
and non-LLM systems differ mainly in what they require of the data, which
Section~\ref{sec:related_work} sets out. A language agent given extracted evidence might fail because it lacks a disciplined procedure for working through the evidence, in which case prompting and agent design should help; or because it lacks the telemetry-specific
knowledge that says which KPI patterns discriminate which fault, in which case generic
procedural prompting alone may not substitute for it.

Procedure and knowledge are routinely varied together. We separate them on
OpenRCA by computing the evidence once, offline, and supplying it
identically to every arm. The OpenRCA agent cannot support that test: its
extraction is part of the reasoning being varied, so the evidence cannot be
held fixed. Our pipeline is therefore the instrument for that comparison: an
offline stage applies fixed per-KPI thresholds to condense each window into a
compact set of anomalous rows, and
a staged pipeline then reasons over that evidence,
calling log- and trace-investigation tools, and the mesh and runtime
sub-agents, on demand. Holding that instrument fixed, we run two controlled experiments, and
evaluate LLM and non-LLM baselines under the same protocol
(Section~\ref{sec:baseline_results}). First, an input-matched ablation
removes the fixed five-step procedure while keeping anomaly inputs, system metadata, tools,
vocabularies, output schema, model identifier and evaluator fixed, which tests the prompt-level
procedure while holding everything supplied alongside it constant. Second, on Market we vary the fault-discrimination rules, supplying none, a manually
authored set, a set mined automatically, or a control written from the fault
labels and KPI schema alone, and transfer the mined rules to a held-out
cloudbed. Both answer a question left open by an error analysis of the agent's
own failures. Given the correct
answer, a reverse reasoning agent reconstructs the evidence chain from the extracted anomalies to
the labelled cause and compares it against the pipeline's own reasoning trace, classifying each failure as a
\emph{Reasoning Gap} (the evidence was present but the system reasoned incorrectly) or
\emph{Data Ambiguity} (no grounded chain was found in the extracted representation). Because the agent is given the label, this is an evidence audit
and not a test of forward identifiability: it locates failures where a grounded
chain existed that the agent did not follow, without saying whether procedure
or knowledge was missing. Those reports are also mined into rules.

Our contributions are:

\begin{itemize}
\itemsep0em
\item An error analysis of an LLM agent's RCA failures on OpenRCA: a
label-conditioned reverse-reasoning audit whose labels agree with two site
reliability engineers, and an onset analysis locating a boundary-anchoring
pattern in the predictions.
\item An input-matched protocol that computes the evidence once, offline, and
supplies it identically to every arm, so procedure and knowledge vary one at a
time. Removing the five-step procedure leaves Full Score higher in all three
conditions we measure, 11 solved cases becoming 22 without rules, though no
contrast survives correction, and deleting one criterion alone gives a net gain
of 8 cases.
\item A rule-mining pipeline that turns label-conditioned reverse-reasoning
audits into grounded, evidence-citing discrimination rules. Mined on Market CB1 and applied to held-out CB2, they
raise Full Score from 14.10 to 35.90, the strongest effect we measure that survives
correction for multiple testing.
\end{itemize}

\section{Related Work}
\label{sec:related_work}

\paragraph{LLM-based systems for OpenRCA.} OpsAgent \citep{luo2025observabilitydatadiagnosisevolving} turns metric, log and trace
anomalies into textual evidence for three cross-reviewing agents, then
optimises their policies in a supervised self-evolution phase that trains on
60\% of the benchmark, which our training-free protocol does not have; it remains
one of the few systems designed for OpenRCA.
GALA \citep{tian2025galagraphaugmentedlargelanguage} combines statistical causal inference with
LLM reasoning, ranking candidates from a metric causal graph with PageRank and
from TWIST span-anomaly scoring on traces before LLM agents re-rank them. Its
accuracy is strongly constrained by the recall of that first ranking, since the
later LLM stages mainly re-rank and elaborate the candidate set.
RCLAgent \citep{zhang2025adaptiverootcauselocalization} models SRE diagnostic workflows
through recursion-of-thought over trace trees, but expects clean single-request
analysis rather than 30-minute windows with concurrent failures, and produces
ranked component lists without reasons or onsets.

All three decompose context reduction, anomaly extraction and
reasoning across agents, which eases the context constraint but increases cost,
leaves results dependent on the base model, and can discard evidence during
reduction.

\paragraph{Recent systems on OpenRCA.} RCAFlow \citep{gao2026rcaflow}
converts troubleshooting documents into behaviour-tree workflows and executes
them through a hierarchical planner, generating analysis code per window. With
DeepSeek-V3 it reports Correct accuracy, the Full Score used here, of 40.44 on
Bank, 34.46 on Market and 47.06 on Telecom, the strongest previously published.
Two differences keep those figures apart: its knowledge is the
fault-handling rule file shipped with the OpenRCA agent, which its baselines
also receive, so its scores sit against our rules ON condition; and its Market
denominator of 148 cases pools both cloudbeds, which the transfer test here
keeps separate (Appendix~\ref{sec:bench_compare}). Its strongest configuration adds SOPs whose construction the paper does not
describe and which, like its code, are not released, where the rules here are mined by a
documented pipeline and released. It also adds procedure where we test whether procedure
helps, and assembles evidence per query rather than once offline, so like the OpenRCA agent it cannot support an
input-matched comparison, and without its knowledge base it falls to
12.50, 10.14 and 15.69 on Bank, Market and Telecom
(Appendix~\ref{sec:bench_compare}). mABC \citep{zhang2024mabc} and Flow-of-Action
\citep{pei2025flowofaction} score root-cause location and type with no onset field and
report no OpenRCA scores of their own; the figures attributed to them are
RCAFlow's reproductions, and neither releases code we could rerun.

\citet{roy2024exploringllmbasedagentsroot} let agents gather diagnostic
evidence on demand, which varies from run to run and so cannot support an
input-matched comparison. \citet{kim2026ai} execute the full OpenRCA benchmark across five models and
classify the observed failures into twelve pitfall types, reporting that the
dominant ones persist across capability tiers and that prompt engineering alone
does not resolve them. Their interventions change how agents communicate and remember within
a window, where the knowledge here is an input that can be withheld or
injected and is tested on a held-out cloudbed.

\paragraph{Non-LLM and causal methods.} Under OpenRCA's roughly 30-sample,
high-dimensional windows both lag-based and conditional-independence
methods are ill-conditioned: at the benchmark's 60-second granularity Granger
causality sees fast effects as simultaneous, and PC yields unstable graphs,
which the extensions PCMCI, FCI, $\Psi$-PC, LiNGAM and RCD relax individual
assumptions of without addressing the data scarcity. We evaluate four of them,
and Section~\ref{sec:baseline_results} finds the outcome dominated by oracle
routing and metric-source choice rather than by the algorithms \citep{pham2025rcaeval}. RUN \citep{lin2024root} extends Granger causality with neural networks and
contrastive learning, but requires training, so we did not evaluate it.
A systematic evaluation \citep{pham2024root} finds no single approach
performs consistently across datasets.

Multimodal non-LLM RCA is established: Nezha
\citep{yu2023nezha}, TORAI \citep{pham2026torai} and MULAN
\citep{zheng2024mulan} need logs and traces as numeric or event features
aligned with the metrics in one representation. OpenRCA's reference agent treats metrics as primary and consults
logs and traces only to verify, and our pipeline follows the same ordering, so
neither builds those features; scoring these methods would need a different
evidence representation (Section~\ref{sec:methods_compared}).
A broad class of methods depends on fault-injection timestamps, normal
baselines, topology, labelled data, or multiple incidents: RCD  \citep{ikram2022rcd}, CloudRanger
\citep{wang2018cloudranger}, EasyRCA \citep{assaad2023easyrca}, MicroDiag
\citep{wu2021microdiag}, MicroCause \citep{meng2020microcause}, MicroRank
\citep{yu2021microrank}, MSCRED \citep{zhang2019mscred}, CausalRCA
\citep{xin2023causalrca}, CIRCA \citep{li2022circa}, CCLH
\citep{xie2025rootcauseanalysismicroservice} and TraceRCA \citep{li2021tracerca}. OpenRCA supplies
none of these directly, and using its evaluation labels as method inputs
would be oracle information. BARO \citep{pham2024baro} and RCAEval's
N-Sigma baseline \citep{pham2025rcaeval} need only an onset and are
evaluated in Section~\ref{sec:baseline_results}. OpenRCA's limited observability rather than its scale is
what distinguishes it: LO2 \citep{lo2} is
larger in raw telemetry volume, RCAEval \citep{pham2025rcaeval} has more
labelled failure cases, and LEMMA-RCA \citep{lemmarca} longer per-incident
series (Appendix~\ref{sec:bench_compare}).

\section{Methodology}

\subsection{System Architecture}

\begin{figure*}[t]
 \centering
 \includegraphics[width=\textwidth]{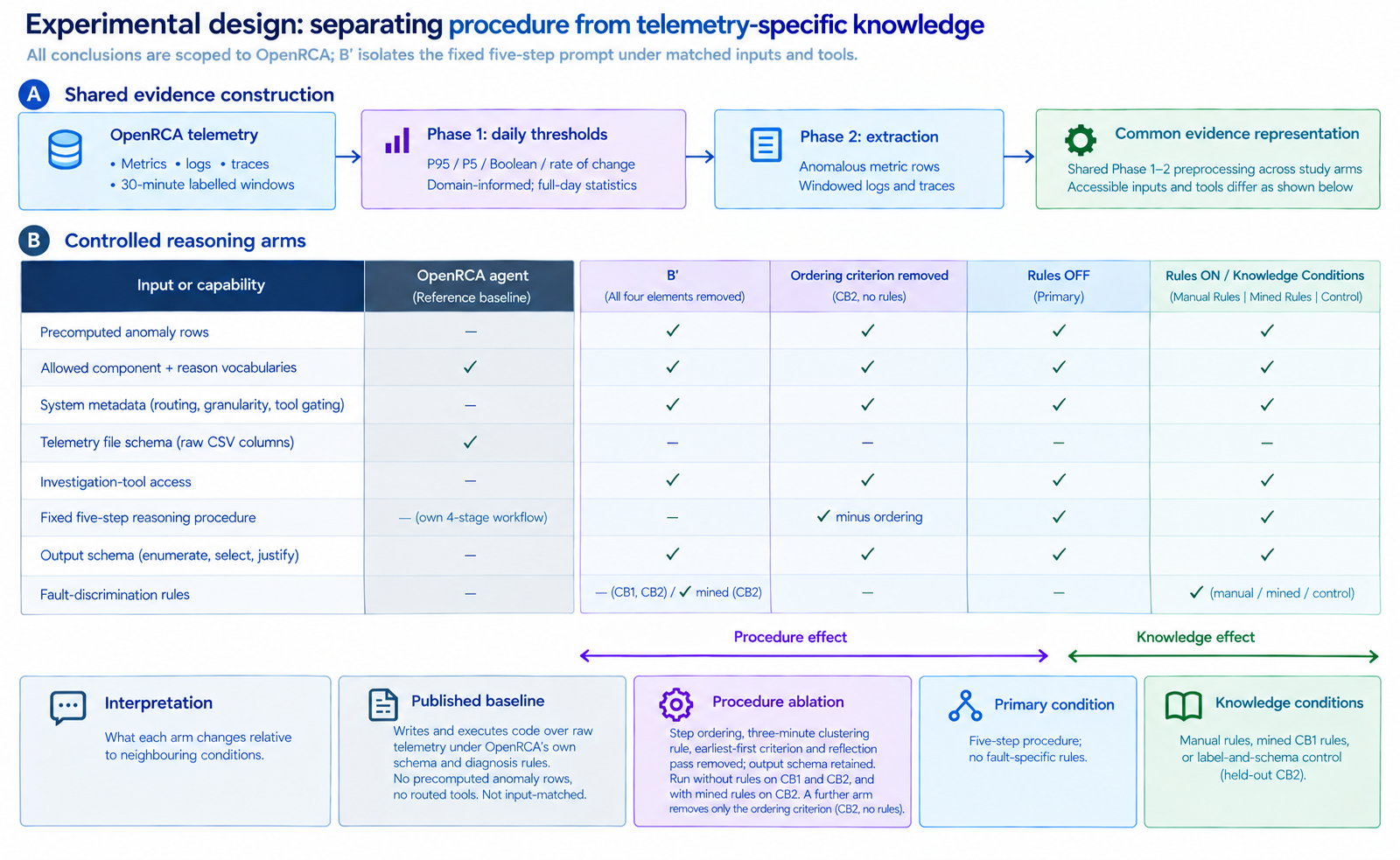}
 \caption{Experimental design. Panel A shows the shared preprocessing that
 constructs the evidence representation. Panel B separates the evaluated reasoning arms and their
inputs: B$'$, which removes all four elements of the procedure, and rules OFF
differ only in the fixed five-step procedure; the single-criterion arm removes
only the onset-ordering criterion; rules OFF and rules ON differ in the
fault-discrimination rules. The OpenRCA agent is shown for reference and is not input-matched
 to the others.}
 \label{fig:architecture}
\end{figure*}

Given the telemetry for a 30-minute failure window, the pipeline predicts a
component name, fault reason and onset timestamp for every case, and the number of root causes in each window (one or two) is
supplied to it; Appendix~\ref{sec:extended_method} states the research questions. It
operates in three phases (Figure~\ref{fig:architecture}).
Phases~1 and~2 prepare data offline; Phase~3 is the reasoning pipeline, the
instrument through which we vary procedure and knowledge independently.

\noindent\textbf{Phase 1: Threshold Pre-computation.}
A preprocessing pipeline analyses each metric source once per dataset and
day, assigns each dynamic KPI one of four anomaly-detection methods (P95, P5,
Boolean or rate-of-change), and computes its threshold from that day's
distribution; KPIs that are effectively constant over the day are treated as
\textit{static} and excluded. An LLM assigns the method from each KPI's summary statistics together with
the fault reasons and component names carried in OpenRCA's own query prompt,
so the assignment is domain-informed but draws on no information the
benchmark withholds from any system. No per-window root-cause labels are used, though the step is transductive,
since each evaluation window contributes to its daily distribution. Thresholding is therefore shared preprocessing, identical across arms.
Appendix~\ref{sec:pseudocode} defines the four methods and the statistics the
assignment uses.

\noindent\textbf{Phase 2: Per-Window Anomaly Extraction.}
Given a 30-minute query window, a scripted extraction step applies the Phase~1 thresholds to each metric source and extracts only the anomalous rows for that window. Log and trace sources are not threshold-processed; instead the full
30-minute window is staged for the investigation tools of
Appendix~\ref{sec:tools}, which the reasoning stage calls on demand.

\noindent\textbf{Phase 3: Reasoning Pipeline.}
Every Stage~1 arm receives the same evidence interface of precomputed anomaly rows
and routed tools, the same \textit{system metadata}, and the same answer
vocabularies (Appendix~\ref{sec:dk_artifacts}). Two further inputs vary: a
fault-discrimination rules document, present under \textbf{rules ON} and absent
under \textbf{rules OFF}, and the reasoning procedure prescribed in the prompt.
Because the metadata itself carries data-source routing, component-granularity and
tool-gating rules, rules OFF is a reduced-knowledge condition rather than a
knowledge-free one, and a knowledge effect reported here is an effect of the rules
document specifically. Three documents are evaluated in that position, the manually
authored, mined and label-and-schema sets of Section~\ref{sec:dk_construction}. The rules contrast is bundled rather than textual, since rules ON also conditions three prompt
passages (Appendix~\ref{sec:pseudocode}). Appendix~\ref{sec:extended_method} describes the evaluated arms in full. Algorithm~\ref{alg:pipeline} states the deployed path end to end.

\noindent\textbf{Stage 1.} Stage~1 is evaluated in three
forms. The first prescribes five steps in one inference pass: \emph{enumerate} every anomalous component, KPI and timestamp;
\emph{cluster} them by component and time, a gap over three minutes starting a new
cluster; \emph{analyse} each cluster for fault type and onset while dismissing
none, applying the injected rules here when present; \emph{select} among the
pooled candidates, keeping any whose onset precedes all others and dismissing a
later one only when it has no independent producer KPI; and \emph{reflect} on the
selection.

The second is B$'$, which replaces those five steps with a direct diagnosis instruction while holding every other input fixed. What it removes is the step ordering, the three-minute clustering rule, the earliest-first selection criterion and the reflection pass, and \emph{procedure} is the short name for those
four throughout. The third removes only the onset-ordering criterion of Step~4
and is reported in Section~\ref{sec:ablation}. The output schema is shared,
so B$'$ must still enumerate its candidates with
evidence, pool them, give selection reasoning and ruled-out
justifications, and label a causal chain;
Appendix~\ref{sec:prompt_stage1} reproduces both prompts in full.

\noindent\textbf{Post-Stage-1 Processing.} One post-Stage-1 step is deployed: a config agent maps
predictions onto the allowed component and reason vocabulary so they are
scoreable. Prior-window screening was evaluated only as an ablation
(Appendix~\ref{sec:prior_window_agents}).

\subsection{Error Analysis Pipeline}
\label{sec:error_analysis}

The error analysis classifies each failure as a Reasoning Gap or
Data Ambiguity. It runs after Stage~1 evaluation and never contributes to the
prediction it analyses. Its reports do feed the rule mining of
Section~\ref{sec:dk_construction}, so mined rules are only ever
evaluated on a cloudbed whose reports they did not use.

\noindent\textbf{Reverse reasoning agent.}
OpenRCA labels only root causes, not which anomalies are causally relevant, so no ground-truth anomaly set exists against which to test reasoning. Instead, a reverse reasoning agent is given the ground-truth answer for each
Stage~1 prediction and reconstructs the reasoning path to it from Stage~1's full
reasoning trace, the complete extracted anomalies for the labelled component,
and the log and trace data on demand. Each step is labelled by knowledge source: \textbf{(A)}
rows Stage~1 retrieved, or \textbf{(A*)} rows present in the anomaly CSVs but
absent from its output, which Stage~1 failed to surface. It runs only in rules
OFF mode, so the path reflects the evidence and not an injected ruleset.

The diagnostic then compares the reconstructed path against Stage~1's
prediction on each dimension: component, reason and timestamp. A dimension is a
\textbf{Reasoning Gap} (Type~1) when the signal was present in (A) or (A*) but
Stage~1 did not reach the correct conclusion, and \textbf{Data Ambiguity}
(Type~2) only when no source the diagnostic can reach holds the signal. Because
the label is supplied throughout, Data Ambiguity is a lower bound: a model given
the answer can build a chain the evidence alone would not single out, and that
case is counted as a Reasoning Gap. Appendix~\ref{sec:worked_cases} gives the
full classification procedure and one worked case per outcome.

\subsection{Domain Knowledge Construction}
\label{sec:dk_construction}

\noindent\textbf{Curated construction.}
Domain knowledge for each dataset was constructed interactively with an LLM coding assistant (Claude Code), by inspecting a representative subset of failure windows against their labels to produce the fault-signature, discrimination, onset and localisation rules injected in rules ON mode.

\noindent\textbf{Automated rule mining.}
To make rule construction more systematic and auditable, we developed a three-phase mining
pipeline over the reverse-reasoning reports. It receives the controlled fault-label vocabulary and the label-conditioned
reports, but no separately authored fault-signature document. The pipeline is therefore label-supervised: it automates extraction, not
supervision. Its one hard constraint is grounding: every rule must cite a
concrete evidence item in the report, a numeric anomaly row or a log or
trace finding, and ungrounded rules are discarded
(Appendix~\ref{sec:mined_rules}, which traces one fault type through all three phases).

\textbf{Phase 1: Mine.} Each reverse reasoning report covers one case, and
therefore one fault reason; it is read together with its Stage~1 output. A mining agent extracts any number of generalisable rules that would have helped Stage~1 reach the ground truth, each free-form text tagged with the ground-truth fault type.

\textbf{Phases 2 and 3: Cluster and Consolidate.} All mined rules are pooled and
clustered by ground-truth fault type, so every window in which a fault occurred
contributes to its cluster whatever Stage~1 predicted. For each fault type a
consolidation agent then merges the pooled rules into one coherent, deduplicated set
spanning recognition, discrimination, localisation and onset, discarding unsupported
statements, and the number of contributing windows is recorded as the rule's support count.

\noindent\textbf{Label-and-schema control.} For comparison we also build a
control document: the same model writes rules for the same fault types from the
fault labels and KPI schema alone, with no anomaly values, no thresholds and no reverse-reasoning reports.

\section{Experiments}

\subsection{Datasets and Evaluation Metrics}
\label{sec:eval_metrics}

\noindent\textbf{Datasets.} All experiments use OpenRCA, the only public benchmark we
found that scores all three fields without supplying the injection time
(Appendix~\ref{sec:bench_compare}): three datasets giving four evaluation domains, since
Market is evaluated on two cloudbeds (Appendix~\ref{sec:dataset_details}). Every system is
given the correct 30-minute window and its associated telemetry; locating the relevant
window is not part of the evaluated task. The 30-minute span is fixed by the OpenRCA protocol,
not a choice we made or tuned.

\noindent\textbf{Metrics.} For LLM-based methods and our system we use the OpenRCA scoring protocol. Each case defines scoring points (component, reason, and onset timestamp, depending on task type) and earns credit per point satisfied, with timestamps correct within 60 seconds, so Full Score means every requested field correct rather than all three. For two-root-cause windows the best assignment is used. A case scores zero if the predicted count differs from the ground truth. We report the OpenRCA
\textbf{Full Score} (the percentage of cases scoring $1.0$) and \textbf{Partial Score} (the
mean per-case score, which credits partially correct predictions). For non-LLM methods we
report \textbf{Accuracy@k} for $k \in \{1,3,5\}$: whether the correct component appears among the
top-$k$ ranked predictions. RCLAgent is scored by Accuracy@1, as it produces ranked component
lists without reasons or timestamps; for GALA we report candidate recall instead.

\subsection{Methods Compared}
\label{sec:methods_compared}

\noindent\textbf{Non-LLM methods.} We evaluate unsupervised causal discovery (PC, FCI,
LiNGAM and Granger) on Market cloudbed-1, which requires none of the
auxiliary information the other methods need, with oracle routing to the
metric source matching the fault's granularity (Appendix~\ref{sec:nonllm_analysis}).

\noindent\textbf{LLM-based methods.} We evaluated the OpenRCA agent on all three datasets (Market, Telecom, and Bank), and RCLAgent on Market cloudbed-1, and GALA on
both Market cloudbeds (Appendix~\ref{sec:gala_failure_analysis}). All systems, including every agent in our pipeline, use GPT-5.2 with the
decoding settings and per-window call structure of
Appendix~\ref{sec:reproducibility}; we additionally re-run the pipeline and the
OpenRCA agent on Market CB1 with Gemma-4-31b-it as a secondary model. We exclude OpsAgent and cite its published results rather than treating
them as a matched baseline (Appendix~\ref{sec:baseline_analysis}).

\section{Results and Analysis}

\begin{table*}[t]
 \centering
 \resizebox{\textwidth}{!}{%
 \begin{tabular}{lcccccccc}
 \toprule
 & \multicolumn{2}{c}{\textbf{Market CB1}} & \multicolumn{2}{c}{\textbf{Market CB2}} & \multicolumn{2}{c}{\textbf{Bank}} & \multicolumn{2}{c}{\textbf{Telecom}} \\
 \cmidrule(lr){2-3} \cmidrule(lr){4-5} \cmidrule(lr){6-7} \cmidrule(lr){8-9}
 \textbf{System} & \textbf{Full} & \textbf{Partial} & \textbf{Full} & \textbf{Partial} & \textbf{Full} & \textbf{Partial} & \textbf{Full} & \textbf{Partial} \\
 \midrule
 OpenRCA agent & 11.43 & 28.69 & 16.67 & 28.42 & 27.94 & 36.64 & 29.41 & 38.22 \\
 \addlinespace
 \textit{Ours, rules OFF (primary condition)} & 17.14 & 35.59 & 14.10 & 28.21 & 27.94 & 37.44 & 35.29 & 44.12 \\
 \quad B$'$: rules OFF minus the fixed five steps & 27.14 & 39.16 & 28.21 & 39.32 & -- & -- & -- & -- \\
 \quad Rules OFF minus the onset-ordering criterion & -- & -- & 24.36 & 42.00 & -- & -- & -- & -- \\
 \addlinespace
 \textit{Ours, rules ON (manually authored)} & 25.71$^{\S}$ & 45.60$^{\S}$ & 24.36 & 45.63 & 36.76$^{\S}$ & 49.70$^{\S}$ & 56.86$^{\S}$ & 60.45$^{\S}$ \\
 \textit{Ours, rules ON (automatically mined)} & -- & -- & 35.90 & 55.03 & -- & -- & -- & -- \\
 \textit{Ours, rules ON (label-and-schema control)} & -- & -- & 19.23 & 31.73 & -- & -- & -- & -- \\
 \quad B$'$ + mined: mined rules minus the fixed five steps & -- & -- & \textbf{42.31} & 54.92 & -- & -- & -- & -- \\
 \bottomrule
 \end{tabular}%
 }
 \caption{Main results on the three OpenRCA datasets, Market reported on both cloudbeds.
 Rows are the OpenRCA baseline agent and our pipeline with rules off (the primary condition), with manually authored rules (rules ON), and
 with automatically mined rules; mined rules are evaluated on held-out CB2 only, where $-$
 denotes not evaluated. B$'$ is the matched-input procedure ablation, run on
Market CB1 without rules and on CB2 both without rules and with the mined
rules. The single-criterion row removes only Step~4's onset-ordering criterion,
on CB2 without rules. The OpenRCA agent is not input-matched
to the others, which share one evidence interface; rules OFF is the arm against
which the others are read. $^{\S}$The Market CB1, Bank and Telecom manual rules were written
against windows from those datasets, so those cells are developed on the data they are scored on;
Table~\ref{tab:heldout} re-evaluates them on construction-excluded subsets.
Scores are OpenRCA Full and Partial. Solved-case counts are in Appendix~\ref{sec:supporting_measurements}. Model: GPT-5.2.}
 \label{tab:system}
\end{table*}

\subsection{Baseline Methods}
\label{sec:baseline_results}

\noindent\textbf{Causal discovery.} Under oracle granularity routing, PC, FCI,
LiNGAM and Granger all fall below a max-anomaly control at Accuracy@3.
Removing the oracle drops that control from 64.3\% to 14.3\%, near chance, so
routing rather than the algorithm dominates. With a label-free onset, N-Sigma is above the
no-oracle control on CB1 and BARO below (Appendix~\ref{sec:nonllm_analysis}).

\noindent\textbf{LLM-based systems.} We report no end-to-end GALA score on Market
cloudbed-1 and instead reconstruct its candidate stage, whose ground-truth
candidate recall is 14/70 (20.0\%); the end-to-end cloudbed-2 run and the
reason for the gap are in Appendix~\ref{sec:gala_failure_analysis}. RCLAgent identifies the labelled component in 8/70
cases (11.43\%) and scores 0/70 Full, since OpenRCA scores a different
combination of fields per query and none of those 8 queries scored on component
alone. Appendices~\ref{sec:nonllm_analysis}
and~\ref{sec:llm_failure_analysis} give the reasons in each case.

\subsection{System Performance}
\label{sec:ablation}

\noindent\textbf{Procedure ablation.} B$'$ exceeds the arm with the procedure in
all three conditions we measure (Table~\ref{tab:system}): $+10.00$ points on
Market CB1, $+14.10$ on CB2 without rules, where 22 of 78 cases are solved against
11, and $+6.41$ on CB2 with the mined rules, where 33 are solved against 28.
Section~\ref{sec:statistical_validity} reports the tests
and Appendix~\ref{sec:statistical_analysis} the intervals. The two gains do not add: an additive prediction gives 39 cases
and the arm solves 33; the mined rules add 17 cases over rules OFF but 11 over
B$'$ (22 to 33).  Overlapping repairs are the largest part of the
shortfall, since of the 14 cases removing the procedure repairs and the
19 the mined rules repair, 7 of the 14 are the same; the two also break 3 and 2 cases, so the
overlap and the six-case shortfall do not coincide exactly. The two also overlap in content: 37 of the 158 mined
rules place the onset or arbitrate between candidates rather than discriminate
between fault types, and seven countermand the earliest-first criterion that
B$'$ removes (Appendix~\ref{sec:rule_content}). B$'$ with the mined rules is
therefore not procedure-free: it replaces the prescribed procedure with the
mined document's onset and arbitration rules, so that contrast compares two
sources of procedure rather than procedure against none.

The arms differ mainly in onset placement. The procedure ranks candidates by
earliest anomaly, treating an earlier onset as evidence against being a
downstream victim, which without fault-specific rules pushes its predictions onto
the 30-minute window boundary where no injected onset can lie: 44\% of rules OFF onsets land there on CB1
against 24\% for B$'$. The benchmark's injection spacing
offers a reason (Appendix~\ref{sec:label_audit}): a window's first samples are often
already anomalous from the preceding fault. The pattern appears only when the procedure is present and rules are absent, with B$'$ matching rules ON at 24\% and the criterion stated identically in both rules modes.

\noindent\textbf{Isolating the criterion.} B$'$ removes four things at
once, so it does not say which of them carries the penalty. We therefore ran a
single-criterion arm on CB2: the onset-ordering criterion of Step~4, which ranks
candidates by earliest onset and keeps the earliest, was deleted and the
three-minute clustering rule, the step ordering and the reflection pass were
left in place. Full Score rises from 14.10 to 24.36, a net gain of 8 cases
against the 11 that removing all four gives, and Partial Score from 28.21 to
42.00. The onset measures move with it: the share of predicted onsets falling
on a window boundary drops from 53\% to 15\%, and the median absolute onset
error from 700\,s to 34\,s, inside OpenRCA's tolerance. The single-criterion arm repairs 11 cases and breaks 3, and 7 of
the 14 cases B$'$ repairs are among its own. The ordering criterion therefore
accounts for most of the procedure's net cost on this cloudbed, though only half
of B$'$'s repaired cases are shared, and for the boundary-onset pattern
specifically. We run no test between this arm and
B$'$ and make no claim that the two are equivalent; the comparison is one run on one cloudbed.

\noindent On CB1, across the answer fields, B$'$ buys onset placement at the
cost of fault naming, though no field-level difference is separable at this size
(Appendix~\ref{sec:statistical_analysis}).

\noindent\textbf{Secondary model.} On Gemma-4-31b-it the gain from rules is
larger than on GPT-5.2 (Appendix~\ref{sec:supporting_measurements}).

\subsection{Domain knowledge and where errors sit}
\label{sec:error_analysis_results}

\noindent\textbf{The evidence audit.}
We use the reverse reasoning of
Section~\ref{sec:error_analysis} as a \emph{label-conditioned evidence audit}. The labelled component had metric anomaly rows in 69 of the
70 cases and rows inside OpenRCA's onset tolerance in 65
(Appendix~\ref{sec:supporting_measurements}), both upper bounds, counting the thresholded metric
sources only, since logs and traces are
staged whole for the tools. It therefore locates reusable discrimination signals for
the mining rather than measuring evidence sufficiency. Two label-aware site reliability engineers agree with the
gap labels on a stratified sample at $\kappa=0.704$ and $\kappa=0.920$, against $\kappa=0.625$ between themselves
(Appendix~\ref{sec:human_validation}).

The Reason dimension has the highest error rate and the largest Type~2
(Data Ambiguity) fraction (Table~\ref{tab:gap_dim}); Appendix~\ref{sec:worked_cases} adds a
paired rules ON comparison.

\noindent\textbf{Mined rules.} Mined rules, learned from CB1
label-conditioned reports and applied to the held-out CB2 cloudbed, whose windows
contribute no reverse-reasoning reports to the mining input, raise Full Score above both rules OFF and the manually authored rules (Table~\ref{tab:system}); Section~\ref{sec:statistical_validity} reports which of these margins separate, and the margin over the manual rules confounds
provenance with supervision volume (Appendix~\ref{sec:dk_artifacts}). Appendix~\ref{sec:mined_rules} reports the coverage and support
distribution of the mined rule set, where Table~\ref{tab:fault_transfer}
decomposes the transfer by fault type.

\noindent\textbf{Grounding.} Because reverse reasoning sees the label, it might build a
chain around evidence that does not discriminate, so every KPI the mined
rules cite is checked against the deployed anomaly representation; a KPI that
never produces an anomaly row cannot reach the pipeline. The mined rules cite such a KPI in 1 of 168
positive citations and never as a primary signal; the control does so in
53 of 180 and selects an absent primary signal for 7 of the 15 fault types
(Table~\ref{tab:grounding}). That audit inspects only the document; run end to end in
the same position on all 63 held-out CB2 windows, the control scores 19.23 against 35.90
(Table~\ref{tab:system}). The control receives the same slot and the same
conditioned passages, so if its own content neither helps nor hurts, $+5.13$
of the $+21.80$ is prompt-side and $+16.67$ document content; its absent-KPI
citations suggest the content may hurt, which would raise the prompt-side share. 

\noindent\textbf{Leakage control.}
\label{sec:leakage_control}
To test the manual rules on windows they were not built
from, we re-evaluate \emph{both} arms on subsets excluding the construction
windows. Only Telecom survives Holm, on 11 discordant cases split 10 to 1
(Table~\ref{tab:heldout}).

\subsection{Statistical Validity} \label{sec:statistical_validity}

\noindent\textbf{What is tested.} Corrections run within families declared post
hoc in Appendix~\ref{sec:statistical_analysis}, never pooled. No procedure contrast
survives: within their own family Holm raises the CB2 no-rules value from
$p=.0248$ to $.0744$, and the single-criterion arm is reported descriptively.

\begin{table}[t]
 \centering
 \footnotesize
 \setlength{\tabcolsep}{4pt}
 \resizebox{\columnwidth}{!}{%
 \begin{tabular}{lrrrr}
 \toprule
 & \multicolumn{2}{c}{\textbf{Full Score}} & \multicolumn{2}{c}{\textbf{Partial Score}} \\
 \cmidrule(lr){2-3} \cmidrule(lr){4-5}
 \textbf{Comparison} & $\Delta$ & \textbf{Holm $p$} & $\Delta$ & \textbf{Holm $p$} \\
 \midrule
 CB2 rules OFF $\rightarrow$ mined & $+21.80$ & \textbf{.0077} & $+26.82$ & \textbf{$<$.001} \\
 Telecom rules OFF $\rightarrow$ manual & $+21.57$ & \textbf{.0444$^{*}$} & $+16.33$ &.0651 \\
 CB2 control $\rightarrow$ mined & $+16.67$ &.0605 & $+23.29$ & \textbf{$<$.001} \\
 Bank rules OFF $\rightarrow$ manual & $+8.82$ &.2276 & $+12.26$ & \textbf{.0096} \\
 CB2 rules OFF $\rightarrow$ manual & $+10.26$ &.2367 & $+17.42$ & \textbf{.0085} \\
 CB2 manual $\rightarrow$ mined & $+11.54$ &.2367 & $+9.40$ &.0837 \\
 CB1 rules OFF $\rightarrow$ manual & $+8.57$ &.2367 & $+10.01$ &.0837 \\
 \bottomrule
 \end{tabular}%
 }
 \caption{Paired comparisons clustered by evaluation
 window and Holm-corrected within each metric over the seven comparisons shown.
 Full Score is the primary outcome; the Partial-Score columns are a secondary
 analysis recomputed after the primary result. Bold marks Holm $p<.05$; Partial values at the
resampling floor are reported as $<$.001. $^{*}$The Telecom Full row is exact McNemar, which that
 domain admits because it carries one case per window; its permutation value is
 $.0067$, Holm $.0402$. Uncorrected values and the resampling settings are in
 Appendix~\ref{sec:statistical_analysis}.}
 \label{tab:paired}
\end{table}

\noindent\textbf{What survives.} Full Score is the primary outcome, and two
comparisons survive Holm on it: the CB1-to-CB2 mined transfer and the
Telecom manual-rules gain (Table~\ref{tab:paired}). All seven are positive. On
Partial Score, a secondary outcome chosen after the primary result, the mined
transfer and the mined-versus-control contrast fall below $.001$ (Appendix~\ref{sec:statistical_analysis}).

\section{Conclusion}

Starting from an error analysis of the agent's failures, then holding the
evidence fixed and varying one ingredient at a time, separates the two
candidate explanations, and the rule document carries the largest gain we
measure. Rules mined from label-conditioned analyses of one cloudbed
transfer to a held-out cloudbed, raising Full Score from 14.10 to 35.90, a gain
that survives correction and the one cross-cloudbed transfer the benchmark
makes available. They score above both hand-written rules and the
label-and-schema control, though neither margin separates on Full Score. In all three conditions we measure,
Full Score is higher without the procedure: without rules it doubles the solved cases, 11 to 22, and deleting only the earliest-onset criterion gives a net gain of 8
cases in one run; with
the mined rules the arm without the
prescribed procedure reaches 42.31, the highest we evaluate on that cloudbed,
though its rules carry onset and arbitration guidance of their own. On both
cloudbeds B$'$ without rules also scores above the manual rules. The two
interventions are sub-additive, with the rules effect the larger in both
comparisons, and the boundary onset placement appears only
when the procedure is present and rules are absent. Mining remains
label-supervised, so what it offers is an auditable route from a set of diagnosed
incidents to reusable fault knowledge.

\section*{Limitations}

\paragraph{Single benchmark and the scope of transfer.} The transfer result
established here is cross-cloudbed: rules mined on CB1 raise Full Score on
held-out CB2, two deployments that share a fault-reason vocabulary, which is
the condition the mined rules are keyed to. OpenRCA is the benchmark this
design needs, and Appendix~\ref{sec:bench_compare} sets out the three properties it
requires; we did not find another that supplies them, so a matched
second-benchmark test is unavailable rather than omitted. The findings
accordingly hold on OpenRCA. Carrying the same protocol to another telemetry
schema, service set or fault taxonomy, and to the reverse CB2-to-CB1 direction
and within-dataset split-half mining, is future work.

\paragraph{Anomaly pre-computation quality.} The fixed per-KPI thresholds of
Phases~1 and~2 could drop evidence before Phase~3 sees it. A threshold sweep
(Appendix~\ref{sec:supporting_measurements}) indicates that the deployed
setting retains the signals the rules rely on, but the cited-KPI sets are
themselves derived from those rules.

\paragraph{Statistical power and single runs.} CB1 supplies 70 labelled cases
over 57 windows and the per-field comparisons rest on 54 or 55 scoring points,
so two comparisons survive correction on Full Score
(Table~\ref{tab:paired}). Differences described as
directional should be read as such. Every score is also a single run, with one exception: the CB2 rules-OFF arm was re-run once and scored 12 of 78
cases against 11 of 78; one case of the 78 changed between the two runs.
One re-run of that arm changed one case of 78, small against the 17-case
contrast it anchors, but budget did not allow repeating the other arms, so
run-to-run variation in the contrasts is unmeasured.

\paragraph{Cost accounting is incomplete.} Per-window call counts are reported
in Appendix~\ref{sec:reproducibility}. We cannot report measured end-to-end
latency, and token usage was not captured for the GPT-5.2 \texttt{main} arms.
The prior-window screening and error-analysis pipelines are diagnostic only.

\paragraph{The diagnostic shares the predictor's model.} Reverse reasoning, gap
classification and rule mining all run on GPT-5.2, the model that produced the
predictions being analysed, so Reasoning Gap and Data Ambiguity
originate with the predictor rather than with an independent judge. Anchoring the
classification on anomaly rows read deterministically from the CSVs limits the
risk, and on a stratified sample of 30 dimensions the labels agree with two
human annotators in the same band as the annotators agree with each other
(Appendix~\ref{sec:human_validation}). The annotators also saw the ground truth, so this
checks evidence sufficiency as the audit defines it, not forward identifiability. The analysis covers the 70 Market CB1
cases in the rules-OFF configuration.

\paragraph{Mechanisms are associational.} The account of why the procedure places onsets on the
window boundary (Section~\ref{sec:ablation}) rests on measured associations
across arms on CB1. On CB2 the ordering criterion was removed directly, so
there the account rests on an intervention. What remains untested on both
cloudbeds is the explanation we offer for why that criterion misleads, namely
that a window's first samples are often left over from the preceding fault. The
sub-additivity of the two interventions is likewise descriptive, a comparison
of observed counts against an additive prediction, with no interaction test.

\bibliography{custom}

\appendix

\section{Extended Methodology}
\label{sec:extended_method}

\noindent\textbf{Research questions.} We seek to address the following research questions. (1)~How do causal-discovery and existing LLM-based RCA systems behave under OpenRCA's short-window protocol? (2)~Holding inputs and tools fixed, what changes when the fixed reasoning procedure is removed, and what changes when fault-discrimination rules are added? (3)~Can label-conditioned failure analyses be converted into reusable rules that improve performance on a held-out cloudbed? (4)~What does a label-conditioned evidence audit reveal about where the pipeline loses information or makes incorrect decisions?

\noindent\textbf{The evidence interface.} The CB1 arms indicate that the evidence interface
matters: the arms that additionally receive precomputed anomaly rows, our system
metadata document and routed tool access are the more accurate ones, although the present comparison does not isolate its individual elements.

\noindent\textbf{The evaluated arms in full.} The arms of Table~\ref{tab:system} are described individually below, with
B$'$ + mined covered under B$'$.

\noindent\textbf{OpenRCA agent.} The baseline agent writes and executes code over the raw telemetry
inside its own reasoning loop, under OpenRCA's own schema block and
diagnosis rules. That block carries the telemetry file descriptions, the
\texttt{cmdb\_id} formats and the same component and reason vocabularies
our arms receive, so what it lacks relative to our pipeline is the
precomputed anomaly tables, our system metadata document and routed tool
access, and it is not input-matched to any of our configurations.

\noindent\textbf{B$'$: input-matched procedure ablation.}
B$'$ receives the same anomaly inputs, system metadata, tools, vocabularies and output schema as
rules OFF, but replaces the fixed five-step prompt with a direct diagnosis instruction. It therefore
tests the prompt-level contribution of the prescribed reasoning procedure while holding fixed the
additional inputs and capabilities bundled with the full pipeline. It was evaluated on Market CB1 without rules, and on held-out CB2 both
without rules and with the mined rules.

\noindent\textbf{Our pipeline, rules OFF.}
The full Phase~3 pipeline operating without injected fault-discrimination rules. Stage~1 receives
system metadata, allowed component and reason vocabularies, and investigation tools, but not the
manually authored or mined fault-signature and localisation rules.

\noindent\textbf{Our pipeline, rules ON (manually authored).}
The full Phase~3 pipeline with fault-discrimination knowledge enabled. Stage~1 additionally
receives domain-specific fault-signature, discrimination, onset and localisation rules. \noindent\textbf{Our pipeline, rules ON (automatically mined).} The rules ON
configuration with the manually authored rules replaced by the automatically
mined rule set of Section~\ref{sec:dk_construction}. It is evaluated on held-out
Market CB2, whose windows contribute no reverse-reasoning reports to the mining
input.

\noindent\textbf{Label-and-schema control.} A rules document written by the same
model from the fault-label vocabulary and KPI schema alone, with no access to
telemetry, labelled windows or reverse-reasoning reports, injected through the
same slot as the mined rules. It isolates what the mined rules contribute beyond
restating the answer space, and is evaluated on Market CB2.

\section{Datasets}
\label{sec:dataset_details}

Table~\ref{tab:dataset_counts} reports the size of each evaluation domain. A \emph{window} is one
30-minute query window; a \emph{case} is one labelled root cause. The two differ because some
windows carry two root causes, so all scores in this paper are computed over cases, not windows.
Held-out analyses use their explicitly reported reduced subsets instead.

The table also gives the size of the answer space, meaning the allowed component and fault-reason
vocabularies a prediction must match exactly. Market's 56 components are 6 nodes, 40 pods and 10
services; Telecom's 43 are 22 nodes, 8 containers and 13 services. Every ground-truth label in all
four domains falls inside its own domain's lists. The spaces differ by a factor of four in
components and three in reasons, so Full Scores are not comparable across domains as a measure of
difficulty: an unguided component-and-reason choice is roughly 1 in 840 on Market against 1 in 112
on Bank.

\begin{table}[tbp]
 \centering
 \small
 \begin{tabular}{lrrrrr}
 \toprule
 \textbf{Domain} & \textbf{Windows} & \textbf{Cases} & \textbf{2-cause}
 & \textbf{Comp.} & \textbf{Reasons} \\
 \midrule
 Market CB1 & 57 & 70 & 13 & 56 & 15 \\
 Market CB2 & 63 & 78 & 15 & 56 & 15 \\
 Bank & 127 & 136 & \phantom{0}9 & 14 & \phantom{0}8 \\
 Telecom & 51 & 51 & \phantom{0}0 & 43 & \phantom{0}5 \\
 \bottomrule
 \end{tabular}
 \caption{Evaluation domain sizes. \textbf{Cases} is the number of labelled root causes and the
 denominator for the full-dataset scores. \textbf{2-cause} counts windows holding two root
 causes. \textbf{Comp.} and \textbf{Reasons} are the allowed vocabulary sizes a prediction must
 match exactly.}
 \label{tab:dataset_counts}
\end{table}

\subsection{What OpenRCA supplies, against other RCA datasets}
\label{sec:bench_compare}

OpenRCA is distinctive in what it withholds rather than in its size.
Table~\ref{tab:bench} sets out what we can source: on labelled case count it is smaller than
RCAEval, and LO2 is far larger in raw telemetry volume. OpenRCA's 335 cases are the sum of
Table~\ref{tab:dataset_counts}; RCAEval's 735 cases over three systems are from
\citet{pham2025rcaeval}; LO2's volume is from \citet{lo2}; LEMMA-RCA spans two domains with two
datasets each \citep{lemmarca}; TelecomTS is a single 5G network \citep{telecomts}. Labelled case counts for LEMMA-RCA and TelecomTS are not directly comparable,
so those cells report scale in the units each paper uses. OpenRCA supplies neither a
fault-injection timestamp nor a verified clean reference period, which is what excludes most of the methods
discussed in Section~\ref{sec:related_work}, BARO and N-Sigma excepted, and forces
onset to be inferred rather than given.

\paragraph{What the comparison requires of a benchmark.} The input-matched
decomposition requires three things: a controlled fault-reason vocabulary that
rules can be keyed to, onset labels scored against a tolerance, and two
deployments sharing one vocabulary, so that rules mined on one can be tested
on the other. Our central manipulation, rules OFF against rules ON, is
undefined without the first, and the cross-cloudbed transfer result is
undefined without the third. OpenRCA supplies all three.
The design also assumes the benchmark supplies the query window and the number
of root causes per window. The query text states that count for all 335 cases,
so it is information the benchmark gives every system rather than an input
only our pipeline receives (Section~\ref{sec:eval_metrics}).

\begin{table}[tbp]
 \centering
 \small
 \begin{tabular}{lll}
 \toprule
 \textbf{Dataset} & \textbf{Scale} & \textbf{Onset / baseline} \\
 \midrule
 OpenRCA & 335 cases, 3 systems & neither supplied \\
 RCAEval & 735 cases, 3 systems & injection time \\
 LEMMA-RCA & 4 sets, 2 domains & longer series \\
 LO2 & ${\sim}2\times10^{9}$ log lines & per-call labels \\
 TelecomTS & $>$1M KPI obs., 5G & -- \\
 \bottomrule
 \end{tabular}
 \caption{What we can source about each dataset; provenance and the blank cells are discussed
 above. Onset / baseline is the column that matters here: OpenRCA supplies neither.}
 \label{tab:bench}
\end{table}

\paragraph{RCAFlow against the scores reported here.} RCAFlow \citep{gao2026rcaflow} evaluates the same three systems. Its own workflow-generation study places Bank between 36.76 and 43.38 across trials. The OpenRCA agent is the one arm both papers run, and
with DeepSeek-V3 it is reported there at 7.35 on Bank, 5.41 on Market and
5.88 on Telecom, against 27.94, 14.19 pooled over both cloudbeds, and 29.41
here, so the shared reference point differs by factors of 2.6 to 5.0. The model identifier differs, GPT-5.2 here against DeepSeek-V3 there, which is the first reason the absolute scores of the two papers are not comparable. Removing RCAFlow's
workflow knowledge base drops it to 12.50 on Bank, 10.14 on Market and 15.69
on Telecom, from 40.44, 34.36 and 47.06 in the same table; its main results table gives 34.46 for Market rather than 34.36. That knowledge base
bundles the diagnostic process, node conditions and tool metadata, and is built
from the procedural rule file distributed with the OpenRCA agent, so the drop
cannot be attributed to fault-discrimination knowledge alone; separating those
factors is what the input-matched design here is for. Its strongest configuration adds unspecified SOPs (43.38, 36.49, 49.02) whose provenance and contents are not reported.

\noindent\textbf{How this work differs from RCAFlow.} RCAFlow is the closest
concurrent system and the strongest previously published result on OpenRCA, so the
difference is worth stating directly. Three things separate the designs. First,
provenance: RCAFlow's knowledge base is the procedural rule file distributed with
the OpenRCA agent, and its strongest configuration adds SOPs whose construction the
paper does not describe and which, like its code, are not released; the rules here
are produced by a documented pipeline from label-conditioned audits, and the rule
document, its per-rule coding and the mining code are released. Second, what varies:
RCAFlow supplies procedure and knowledge together, a behaviour-tree workflow executed
by a hierarchical planner alongside the rule file, so a gain cannot be attributed to
either; the design here holds the evidence fixed and varies them one at a time. Third,
evidence assembly: RCAFlow generates analysis code per window, so what the reasoner
sees depends on the arm, whereas the evidence here is computed once offline and
supplied identically to every arm. We make no claim to a higher score: the model
identifiers differ and RCAFlow is above our Bank cell. The comparison is confined to
OpenRCA, the only benchmark we found that supplies the three properties of
Appendix~\ref{sec:bench_compare}, and RCAFlow reports its OpenRCA results on the same
three datasets.

\subsection{Structural audit of the Market CB1 labels and windows}
\label{sec:label_audit}

Low absolute scores can reflect the benchmark as well as the method, so we
audited the structure of all 70 Market CB1 cases
(\texttt{code/label\_audit\_summary.py}). This is a structural audit only: it checks how the labels
and windows are laid out, and does \emph{not} check whether a label is
semantically correct.
Three properties bear on how the scores should be read (Table~\ref{tab:label_audit}). First, 26
of the 70 cases share a window with a second injected fault, and the median interval between
injections, 26.6 minutes, is shorter than the 30-minute window itself, with 41 of 69 gaps below
it and the shortest at 3.1 minutes. Evidence from an adjacent fault is therefore present inside
many windows by construction. Second, no injection lands on the 60-second sampling grid, so the
labelled onset always falls between two samples, a median of 14 seconds from the nearer one;
against OpenRCA's $\pm$60-second tolerance this is not disqualifying, but it means an exactly
correct onset is not observable. Third, all 25 service-level cases carry a container-level fault
reason, so the label mixes granularities within a single case. Separately, one label uses the service name
\texttt{frontend}, whereas the telemetry sources in this cloudbed use instance-level identifiers
such as \texttt{frontend-0}. Resolving that case therefore requires aliasing or service-level
aggregation rather than a direct name match.

\begin{table}[tbp]
 \centering
 \footnotesize
 \begin{tabular}{lr}
 \toprule
 \textbf{Property} & \textbf{Value} \\
 \midrule
 Labelled cases & 70 \\
 Evaluation windows & 57 \\
 \quad with one labelled fault & 44 \\
 \quad with two & 13 \\
 Cases sharing a window with another & 26 \\
 \addlinespace
 Consecutive injection gaps & 69 \\
 \quad median gap & 26.6 min \\
 \quad shorter than the 30-min window & 41 (59\%) \\
 \quad shortest & 3.1 min \\
 \addlinespace
 Injections on the 60-s sampling grid & 0 of 70 \\
 \quad median distance to nearest sample & 14 s \\
 \addlinespace
 Cases labelled at node / service / pod level & 26 / 25 / 19 \\
 \quad service-level with a container-level reason & 25 of 25 \\
 \bottomrule
 \end{tabular}
 \caption{Structural audit of Market CB1. Distance to the nearest sample is
 $\min(t \bmod 60,\; 60 - t \bmod 60)$; the offset past the preceding sample has a median of
 30\,s instead. No claim is made here about semantic label correctness.}
 \label{tab:label_audit}
\end{table}

\clearpage

\noindent\textbf{Domain telemetry.} Market is modelled after an Online
Boutique-style microservice deployment, evaluated on two independent
cloudbed configurations (CB1 and CB2) that share the same topology and
fault vocabulary but differ in component instances and failure windows;
its telemetry spans five metric source types (container, mesh, node,
runtime, service), distributed traces, service logs and proxy logs, and
it requires multi-level reasoning across container, service and node
hierarchies. Bank provides container and application metrics, service
logs and trace spans. Telecom provides container, node, application,
middleware and service metrics and trace spans, with no log data.

\section{Baseline Method Analysis}
\label{sec:baseline_analysis}
We distinguish methods we did not evaluate
from methods that are inapplicable on this benchmark.

\noindent\textbf{Methods reported from published numbers or as diagnostics.} One comparison is not run as a matched arm. OpsAgent is excluded under a stated rule: a
published method whose core learning phase cannot run in this protocol is excluded
rather than approximated.

\noindent\textbf{OpsAgent's published numbers.} Averaged over Telecom, Bank
and Market, OpsAgent reports 16.54\% Correct and 41.35\% Partial with a
Qwen2.5-14B-Instruct-1M seed model, against 11.28\% and 32.33\% for RCA-Agent
with Claude 3.5 Sonnet \citep{luo2025observabilitydatadiagnosisevolving}; its
per-system Correct scores are 30.00, 18.52 and 10.17. Those figures are
obtained after training on 60\% of the benchmark and are computed on the
remaining 40\% as a test split, and their Partial counts queries with at least
one subtask correct rather than OpenRCA's fractional score, so they share
neither a denominator nor a metric with Table~\ref{tab:system}.
BARO and N-Sigma are run with a label-free onset
(Appendix~\ref{sec:nonllm_analysis}); the packaged implementations take that onset
as an argument.

Tables~\ref{tab:nonllm} and~\ref{tab:llmmethods} report the baseline scores discussed in Section~\ref{sec:baseline_results}.

\begin{table}[t]
 \centering
 \small
 \begin{tabular}{lccc}
 \toprule
 \textbf{Method} & \textbf{A@1} & \textbf{A@3} & \textbf{A@5} \\
 \midrule
 Max-anomaly control$^{\S}$ & 25.7 & \textbf{64.3} & 82.9 \\
 \addlinespace
 \texttt{fci\_pagerank} & 18.6 & 52.9 & 67.1 \\
 \texttt{pc\_pagerank} & 20.0 & 47.1 & 58.6 \\
 \texttt{granger\_pagerank} & 20.0 & 45.7 & 58.6 \\
 \texttt{lingam\_pagerank}$^{\ddagger}$ & 21.4 & 40.0 & 52.9 \\
 \texttt{*\_randomwalk}$^{\dagger}$ & 25.7 & 41.4 & 52.9 \\
 \texttt{fci\_randomwalk} & N/A & N/A & N/A \\
 \addlinespace
 Chance (coverage-weighted)$^{\P}$ & 11.5 & 34.5 & 54.6 \\
 \bottomrule
 \end{tabular}
 \caption{Non-LLM causal discovery on Market cloudbed-1, oracle-routed to the metric
 source matching ground-truth granularity, over 70 cases. Failed or missing outputs are counted
 as incorrect, except \texttt{fci\_randomwalk}, which is marked not applicable because its
 scoring head cannot process FCI's PAG edge marks. Every applicable method falls below the
 max-anomaly control at A@3. $^{\S}$The control ranks components by the largest robust deviation
 among the same selected variables, with no graph and no PageRank; it is a within-window
 statistic, not a two-sample comparison against a normal reference. $^{\dagger}$The PC, Granger and LiNGAM random-walk variants are one
 row because RCAEval's random-walk head ignores the adjacency matrix, so they do not measure
 those algorithms. $^{\ddagger}$LiNGAM is non-deterministic: pooled over ten draws from two
 independent five-repetition runs, A@3 has mean 38.3 and standard deviation 2.51.
 \texttt{fci\_randomwalk} fails on FCI's PAG edge marks. Removing oracle routing while holding
 three metric sources fixed moves the control to 14.3 at A@3, approximately the corresponding
 random expectation, and does not preserve the ordering above. $^{\P}$Chance is coverage-weighted over the candidate sets each method produces; the formula, the set sizes and the A@10 discussion are in Appendix~\ref{sec:nonllm_analysis}.}
 \label{tab:nonllm}
\end{table}

\begin{table}[tbp]
 \centering
 \small
 \setlength{\tabcolsep}{4pt}
 \resizebox{\columnwidth}{!}{%
 \begin{tabular}{lccc}
 \toprule
 \textbf{Method} & \textbf{Cand. recall} & \textbf{Comp. A@1} & \textbf{Full} \\
 \midrule
 GALA (candidate stage) & 14/70 = 20.0 & -- & -- \\
 RCLAgent$^{\P}$ & -- & 8/70 = 11.43 & 0/70 \\
 \bottomrule
 \end{tabular}%
 }
 \caption{LLM-based multi-agent methods on OpenRCA Market cloudbed-1. The candidate-stage reconstruction and the RCLAgent protocol are described in
 Appendices~\ref{sec:gala_failure_analysis} and~\ref{sec:llm_failure_analysis}. $^{\P}$Scored on GPT-5.2 transcripts only, over all 70 labelled cases.}
 \label{tab:llmmethods}
\end{table}

\subsection{Non-LLM Causal Discovery: Protocol and Representation Constraints}
\label{sec:nonllm_analysis}

\noindent\textbf{Protocol.} We preserve component identity rather than averaging
KPIs across entities, exclude timestamps from the variable set, and count every
crash and missing output as incorrect. We report two references: a coverage-weighted chance level and a max-anomaly control that ranks components by the largest robust deviation among the same selected variables, with no graph and no PageRank. Chance is coverage-weighted, $\frac{1}{N}\sum_i \mathbb{1}[y_i \in C_i]\min(k/|C_i|, 1)$, computed over the same candidate sets each method produces, whose sizes have median 11 and range 3 to 18. Two arms fall below chance at A@5, \texttt{lingam\_pagerank} and the random-walk row, both at 52.9 against 54.6. We do not report A@10 because 286 of 630 candidate sets hold ten or fewer items, so the metric is near-vacuous here; on these set sizes chance at A@10 is 77.6, not the ${\sim}18$ implied by a 56-candidate assumption.

\noindent\textbf{Methods needing only an onset.} BARO \citep{pham2024baro}
and the N-Sigma baseline distributed with RCAEval \citep{pham2025rcaeval} require
an injection time rather than a normal baseline or a topology, so an onset
estimate is enough to run them. RCAEval's packaged implementations take that time as an
argument and do not estimate it, unlike published BARO; we restore that
step below. We use the window midpoint, which needs no labels. OpenRCA
builds each query window around an injection, so the midpoint is nearly unbiased
but imprecise: the median signed error is $+19$\,s on CB1 and $-99$\,s on CB2,
while the median absolute error is 405\,s and 446\,s, with a worst case near
15 minutes. Its fit to the labelled onsets is reported here, not tuned on
them. As an onset predictor in its own right the midpoint is weak: it falls
within OpenRCA's $\pm$60\,s tolerance for 8 of 70 CB1 cases (11.4\%) and 5 of 78 on CB2 (6.4\%), against
38\% for rules OFF and 53\% for both B$'$ and rules ON, the latter two over the
onset scoring points rather than all cases.
Rules OFF beats the midpoint within tolerance, 38\% against
11.4\%, so the onset field does carry information; but because 44\% of its
onsets sit on the window boundary its median absolute error, 468\,s, stays
close to the midpoint's 405\,s, whereas B$'$ and rules ON reach 49\,s and
44\,s. Both run on
the same three metric sources as the max-anomaly control and without oracle
routing, so the three are directly comparable. N-Sigma reaches 10.0\%
Accuracy@1 and 24.3\% Accuracy@3 on CB1 and 9.0\% and 23.1\% on CB2; BARO
reaches 2.9\% and 11.4\% on CB1 and 3.8\% and 12.8\% on CB2. The max-anomaly control was computed on CB1 only, so
the comparison is made there: N-Sigma is above its 14.3\% at Accuracy@3 and BARO
below it. N-Sigma also has the strongest Accuracy@1 of any non-LLM method we run
without oracle routing. Supplying the labelled
onset instead raises BARO to 20.0\% and 25.6\% and N-Sigma to 30.0\% and
32.1\% at Accuracy@3, but that uses the answer's third field and is
reported only as headroom. The midpoint's signed error is small because errors cancel in both
directions; its typical absolute error is about seven minutes, and it lands
within two minutes of the labelled onset for only 14 of 70 CB1 cases and 12 of
78 on CB2.

\noindent\textbf{BARO with its own change point.} Published BARO detects the
change point itself, which is the step RCAEval's implementation omits, so the
onset above is supplied from outside. To separate the method from that
omission we restored the step: one change point per window, found by dynamic
programming over the window's own robust-scaled aggregate (\texttt{ruptures}, $L_2$
cost, a single breakpoint, no labels), passed to BARO as a row index in place of
a timestamp. This is not the Bayesian online detection of the published
method, but it removes the external onset. It raises BARO's Accuracy@3 from 11.4\% to 12.9\% on CB1,
still below the 14.3\% control there, and from 12.8\% to 15.4\% on CB2.

The detected point is not close to the labelled onset. Its median absolute
error is 496\,s on CB1 and 577\,s on CB2, against the midpoint's 405\,s and
446\,s, and it falls inside the $\pm$60\,s tolerance for 8 of 70 and 6 of 78
cases. It ranks about as well as the midpoint, slightly higher on
both cloudbeds, though by one case of 70 and two of 78, which is too little to
read much into. One possibility is that what BARO needs is a split separating
normal from anomalous behaviour rather than the onset itself. The labelled
onset is better still, at 20.0\% and 25.6\%.

\noindent\textbf{Per-method caveats.} The ordering of PC, FCI, Granger and LiNGAM is not stable across source sets, so we do not rank them. Datasets that supply substantially longer per-incident series, such as LEMMA-RCA \citep{lemmarca}, would be needed to separate the effect of the short window from the
methods themselves.

Our evaluation is restricted to metric-only implementations operating on flat numeric matrices; it
therefore does not test multimodal non-LLM RCA methods. OpenRCA additionally represents root
causes at container, service and node levels, whereas the evaluated implementations rank variables
from one selected metric source. These are constraints of the tested protocol, not general
limitations of causal discovery.

Each method is run on the metric source matching the ground-truth granularity of the fault, on
wide-format matrices built from the raw CSVs. For the container source, for example, the original approximately 2500 metric columns were reduced to 640 through preprocessing: low-variance or constant metrics across the 30 timestamps were removed, redundant time-related features were eliminated, and columns containing only NaN or Inf values were discarded.

\subsection{GALA}
\label{sec:gala_failure_analysis}

\noindent\textbf{Which graph is the hard part.} Where the model builds graphs,
as in GALA, the service dependency graph is not the hard part, since it follows
deterministically from parent-child trace relations and is routinely built at
production scale; the difficulty is the causal graph over KPI variables, whose
estimated edges and heuristic weights are unreliable without a ground-truth
structure.

\noindent\textbf{Coverage.} GALA was evaluated fully on CB2 and only through
its first phase on CB1.

\noindent\textbf{Reconstruction.} Our candidate stage implements a metric ranking by \emph{Granger}
causality with PageRank scoring. Metric selection keeps the top 50 variables by variance, and the
candidate list is truncated at the released caps. The trace side implements TWIST-style span
anomaly scoring. Each query was given the correct 30-minute window and its container, trace and
proxy-log slices.

The per-pod bundling and the Phase-3 agent chain (re-ranking, deep dive,
remediation) are described here but not scored.

\noindent\textbf{What the candidate stage constrains.} On the reconstructed CB1 stage, recall is 14/70 (20.0\%), namely 14/19 pod cases and 0/51 node
or service cases, because the stage emits pod-named candidates. On the CB2 run recall is 48.7\% (38/78), and 5 of those 38 candidate-present cases become correct final answers. Candidate
recall is a strong bottleneck but not a strict upper bound in this implementation: 2 further cases
were finally correct although the labelled component was absent from the recorded candidate list
(\texttt{node-6} and \texttt{frontend}), giving 7/78 = 9.0\% component accuracy overall.

\noindent\textbf{Output format.} Even where the candidate is present, the output is a narrative
incident report rather than a $\langle$component, reason, timestamp$\rangle$ prediction. A
representative report names four services in priority order (\texttt{frontend-2},
\texttt{checkoutservice2-0}, \texttt{cartservice-2}, \texttt{currencyservice-0}) and attributes
them to generic conditions, resource saturation, HTTP redirect misconfiguration and inefficient
processing, none of which belongs to the OpenRCA reason vocabulary, and it supplies no onset
timestamp. A stronger first stage would therefore not address a stage that misses
two thirds of the label hierarchy by construction.

\subsection{RCLAgent}
\label{sec:llm_failure_analysis}

\noindent\textbf{Protocol and failure modes.} RCLAgent is scored on GPT-5.2 only, matching the protocol used for every other system here, over all 70 labelled cases, with any case lacking a scoreable output counted as incorrect. Under first-run selection it relies primarily on trace-tree traversal, which does not directly observe infrastructure-node failures, outputs service-level component names that fail exact-match scoring against pod-level ground truth, and supplies no onset, so no case satisfies all of its scoring points.

RCLAgent selects the single most anomalous trace by duration-to-baseline ratio, but the longest-duration trace often does not correspond to the labelled root cause. Many OpenRCA ground-truth answers are node-level (\texttt{node-2},
\texttt{node-4}, \texttt{node-6}), and infrastructure nodes do not
generate trace spans. For pod-level cases, ground truth specifies identifiers such as \texttt{recommendationservice2-0}, while RCLAgent often outputs service-level names such as \texttt{productcatalogservice}; under exact-match scoring these are incorrect. To separate coarseness from error we rescore the same GPT-5.2 transcripts with both prediction and label collapsed to the service, stripping any trailing pod index. RCLAgent's service-level component accuracy rises from 8/70 (11.43\%) to 14 of 70 (20.00\%), so six cases name the right service but the wrong pod, and the remaining cases are wrong at every granularity. Full Score stays 0/70: OpenRCA scores a different
combination of fields per query, none of the component-correct cases scored on
component alone, and collapsing component names cannot supply the reason or the
onset RCLAgent never predicts. Granularity therefore accounts for part of the component gap and none of the Full-Score gap.

One input asymmetry should be read alongside this. Our pipeline is told the
pod-versus-service decision explicitly: the injected system metadata
(Appendix~\ref{sec:dk_artifacts}) states that \texttt{cmdb\_id} in the container
metrics is formatted \texttt{<node>.<pod>} and must have its node prefix
stripped, that a single affected pod is reported by its own name, and that a
service name is reported when three or more of a service's four pods are
affected simultaneously. That metadata is present in every configuration of our
pipeline, rules OFF included, whereas RCLAgent receives no equivalent
instruction. Part of the component gap in Table~\ref{tab:llmmethods} therefore
reflects that input rather than reasoning alone, which is what the
service-level rescore isolates. RCLAgent's predictions also cluster on generic reasons such as \textit{container network latency}, whereas the ground truth contains diverse fault types including read I/O load, packet corruption, and node CPU spike.

\section{Mined Discrimination Rules}
\label{sec:mined_rules}

\begin{table}[tb]
 \centering
 \small
 \setlength{\tabcolsep}{4pt}
 \resizebox{\columnwidth}{!}{%
 \begin{tabular}{lcc}
 \toprule
 & \textbf{Mined} & \textbf{Label-only} \\
 \midrule
 \multicolumn{3}{l}{\emph{Document grounding, CB1}} \\
 Absent primary signal & \textbf{0/15} & 7/15 \\
 Citations of absent KPIs & \textbf{1/168} & 53/180 \\
 \quad as a percentage & \textbf{1\%} & 29\% \\
 \midrule
 \multicolumn{3}{l}{\emph{End to end, held-out CB2}} \\
 Full Score & \textbf{35.90} & 19.23 \\
 Partial Score & \textbf{55.03} & 31.73 \\
 Solved only by this arm & \textbf{16} & 3 \\
 \bottomrule
 \end{tabular}%
 }
 \caption{Mined rules against a label-and-schema control written by the same model from the fault
 labels and KPI schema alone. Full Scores are $28/78$ and $15/78$. Model: GPT-5.2.}
 \label{tab:grounding}
\end{table}

Mining over the 70 Market CB1 reverse-reasoning reports produced 429 evidence-citing candidate rules (mean 6.1 per case), consolidated into 158 rules spanning all 15 CB1 fault types. Support counts sum to 70, so every analysed case contributes to exactly one fault type's rule set. Support is uneven: it ranges from 8 cases (container read I/O load, the worked example below) down to 1 (container write I/O load), so the rarest fault types are generalised from very few observations. Table~\ref{tab:fault_transfer} sets that support against what transferred.
Rules are not monotonically helpful: Appendix~\ref{sec:dk_hurts} collects the
cases they break. The consolidated document is the artifact injected as mined domain knowledge and evaluated on held-out Market CB2 (Table~\ref{tab:system}); the per-fault-type rule sets, with their support counts and cited evidence KPIs, are included in the supplementary material under \texttt{rule\_mining/mined\_rules/}. Citations are classified by role, so that a rule saying \emph{do not use X} does not count as recommending X.

\begin{table}[tbp]
\centering\footnotesize
\resizebox{\columnwidth}{!}{%
\begin{tabular}{@{}lrrrrrr@{}}
\toprule
& \multicolumn{2}{c}{CB1 mining} & \multicolumn{4}{c}{Held-out CB2 reason} \\
\cmidrule(lr){2-3}\cmidrule(lr){4-7}
Fault type & sup. & rules & $n$ & OFF & cur. & mined \\
\midrule
container network packet retransmission & 5 & 15 & 9 & 0 & 0 & 0 \\
container memory load & 7 & 11 & 7 & 4 & \textbf{5} & \textbf{5} \\
container CPU load & 7 & 13 & 6 & \textbf{6} & \textbf{6} & 5 \\
container network packet corruption & 6 & 12 & 6 & 1 & 4 & \textbf{5} \\
container read I/O load & 8 & 11 & 6 & 0 & \textbf{6} & \textbf{6} \\
node CPU load & 4 & 9 & 6 & 2 & 1 & \textbf{5} \\
container packet loss & 2 & 7 & 5 & 0 & \textbf{3} & 0 \\
node memory consumption & 5 & 12 & 4 & 0 & 0 & \textbf{1} \\
container network latency & 4 & 10 & 3 & 0 & 1 & \textbf{2} \\
node disk write I/O consumption & 5 & 10 & 3 & 0 & \textbf{1} & \textbf{1} \\
container process termination & 4 & 12 & 2 & 0 & 0 & 0 \\
container write I/O load & 1 & 6 & 2 & \textbf{1} & 0 & \textbf{1} \\
node disk read I/O consumption & 5 & 10 & 2 & 1 & \textbf{2} & \textbf{2} \\
node disk space consumption & 5 & 12 & 2 & 0 & 0 & 0 \\
node CPU spike & 2 & 8 & 1 & 0 & 0 & 0 \\
\midrule
All points & 70 & 158 & 64 & 15 & 29 & \textbf{33} \\
Per case & & & 48 & 22.9\% & 42.7\% & \textbf{49.0\%} \\
\bottomrule
\end{tabular}%
}
\caption{Per-fault-type transfer of the CB1-mined rules to held-out Market
CB2. \emph{sup.} is the number of CB1 cases a fault type was mined from and
\emph{rules} the number of rules consolidated for it; the columns sum to the
70 analysed CB1 cases and the 158 rules of Appendix~\ref{sec:mined_rules},
and every one of the 15 CB2 fault types has a mined rule set. The CB2
columns count correctly named fault types among the 64 distinct reason
scoring points carried by the 48 reason-asking CB2 cases, under rules OFF,
the curated rules, and the mined rules. A reason scoring point is the
ground-truth fault-type string itself, so no fault has to be attributed to
a position inside a multi-fault window; two same-type points in one window
are counted once, which is the resolution the stored scoring has. The
\emph{Per case} row averages points within a case. Mined rules lead on 8 of the 15 types against rules
OFF, tie on 6 and trail on 1, and they move four types off zero; against
the curated rules the split is 5, 8 and 2.
The least-supported types do not transfer: the three with
support $\leq 2$ stay at 12.5\% under both, while the twelve with support $\geq 3$
rise from 25.0\% to 57.1\%, which is the cost of generalising a fault type
from one or two observations. This decomposition is descriptive.
Per-type $n$ is 1 to 9, and these comparisons are not among the four declared families of Appendix~\ref{sec:statistical_analysis}, so no corrected
verdict is claimed for any row.}
\label{tab:fault_transfer}
\end{table}

``Evidence-citing'' means that the generating agent attached a
concrete anomaly, log or trace item; it is not an independent validation that the cited item is
sufficient for forward discrimination. The evidence is read from the anomaly CSVs rather
than from Stage~1's output. Because the report's anomaly rows are pre-read deterministically from the
ground-truth component's raw CSVs, they reflect that component's
signature even when Stage~1 localised wrongly. The mined rules are therefore
anchored on the labelled component's own rows, which favours discrimination between fault types over
discrimination against competing components. Rules are kept by grounding
rather than by the gap type the reverse agent assigned, because skipping Type~2
windows would discard candidate rules for faults with only an indirect or
combined signature.

\noindent\textbf{Where the control fails.} The control's errors are concentrated rather than uniform, which is what the grounding audit predicts. OpenRCA tasks ask for different field subsets: 33 of the 78 CB2 cases require one field and only 6 require all three. The control reaches 45.5\% where only a component is asked, and $0/11$ where an onset and a reason are both required, 10.3\% across the two-field tasks. It can localise and often time the onset; what it cannot do is name the fault, which is the discrimination the mined rules supply.

We trace one fault type, \textit{container read I/O load} (\texttt{C1}) in Market CB1,
through the three phases of the mining pipeline described in
Section~\ref{sec:dk_construction}.

\paragraph{Phase 1: Mine.}
Eight CB1 cases carry \texttt{C1} as their ground-truth reason. Each is mined independently into
candidate rules, on average six per window, and every candidate must cite concrete evidence rows
or be discarded. One candidate mined from window \texttt{2022\_03\_20/09-00\_09-30}
(\texttt{shippingservice-1}) reads:

\begin{quote}\small
\emph{When filesystem read KPIs} (\texttt{container\_fs\_reads*} \emph{/}
\texttt{container\_fs\_reads\_MB*}) \emph{show an abrupt, extreme spike at incident onset on a pod,
classify the primary reason as container read I/O load even if CPU usage and CFS throttling are
also anomalous on the same pod.}
\end{quote}

\noindent It cites four rows from that window's anomaly data, three of which are:

\begin{quote}\scriptsize\ttfamily
1647738540 | container\_fs\_reads\_MB./dev/vda\\
\hspace*{1.2em}= 4233.5 (thr 0.0, P95)\\
1647738600 | container\_fs\_reads\_MB./dev/vda\\
\hspace*{1.2em}= 15138.9 (thr 0.0, P95)\\
1647738540 | container\_cpu\_cfs\_throttled\_seconds\\
\hspace*{1.2em}= 3.30 (thr 0.0, P95)
\end{quote}

\noindent The same window yields the discrimination this fault type needs: \emph{treat large
paging/major-fault signals co-occurring with high filesystem reads as supporting evidence of read
I/O pressure (page-ins), and avoid labelling the incident as purely CPU-throttling when these
memory-fault indicators appear at the same onset time}.

\paragraph{Phase 2: Cluster.}
Candidates from all eight windows, whose labelled components mix
pod-level and service-level names, are pooled on the ground-truth fault
type, irrespective of what Stage~1 predicted in each window.

\paragraph{Phase 3: Consolidate.}
The pool is merged into a single rule set for \texttt{C1}: \textbf{11 rules citing 13 distinct
evidence KPIs}, with support~8. Its recognition signature is an abrupt, sustained spike in
\texttt{container\_fs\_reads}/\texttt{\_MB} from a near-zero baseline to multi-GB read volume or
more than 10,000 read operations per scrape near onset (observed volume range
${\sim}2410$--$15138$\,MB), optionally corroborated by node
\texttt{system.io.r\_await} or \texttt{system.cpu.iowait} elevation, with co-occurring CPU or
memory spikes treated as secondary kernel I/O work.

\paragraph{Curated and mined rules for C1.}
The curated Market document classifies C1 by an absolute magnitude:
\texttt{fs\_reads\_MB} above 100~MB per scrape interval, confirmed by sustained pgfault. The
mined rule set states no magnitude. It requires that \texttt{container\_fs\_reads} and
\texttt{container\_fs\_reads\_MB} jump from a near-zero baseline to very large values and remain
anomalous across consecutive timestamps, and that read throughput and read operations rise
together before a read-I/O mechanism is preferred over a CPU-only, memory-only or write-only
explanation. The mined document also names almost no Market entities. A case-insensitive
search for CB1's 21 distinct component identifiers, its six nodes, nine services
and six pod-instance forms, matches only two of them: \texttt{node-6} twice, and
both cloudbeds contain all six nodes, so this is shared topology rather than a
CB1-specific name; and \texttt{emailservice} once, which is never a labelled root
cause on CB2 and so cannot win a case there. The rules are otherwise stated over
KPI patterns. The mined form is relative where the curated form is absolute. A 100~MB constant is
fixed to the cloudbed it was written against, whereas a near-zero-baseline criterion carries to a
cloudbed with different instances, which may be one reason mined knowledge transfers to CB2
(Table~\ref{tab:system}).

Several of these signatures could be anchored only on (A*) evidence, namely numeric read rows
present in the anomaly data but surfaced only as prose in the Stage~1 output.
Appendix~\ref{sec:case_astar} works through one such window in detail.

\subsection{What the mined rules contain}
\label{sec:rule_content}

The mined document occupies the same prompt slot as the manual and control
documents, but its content is not purely fault-discriminative. We coded all
158 consolidated rules across the 15 Market fault types into four categories
(Table~\ref{tab:rule_content}): \emph{discrimination}, which KPI pattern
indicates which fault, including differential diagnosis between fault types;
\emph{localisation}, which component or granularity to attribute a fault to;
\emph{onset placement}, where to put the onset timestamp; and \emph{candidate
arbitration}, how to choose among competing candidates in a window. The last
two are procedural: they do the work Step~4 of the five-step procedure does.
Discrimination and localisation account for 121 rules (76.6\%), onset placement
and arbitration for 37 (23.4\%). The coding was done by one of the authors. There is no
second coder and no inter-coder agreement statistic, so the per-rule coding is
released with the rule document and the classification can be checked against
it.

\begin{table}[tb]
\centering\small
\begin{tabular}{lrr}
\toprule
\textbf{Category} & \textbf{Rules} & \textbf{Share} \\
\midrule
Discrimination & 105 & 66.5\% \\
Localisation & \phantom{0}16 & 10.1\% \\
Onset placement & \phantom{0}23 & 14.6\% \\
Candidate arbitration & \phantom{0}14 & \phantom{0}8.9\% \\
\midrule
\emph{Fault knowledge} & 121 & 76.6\% \\
\emph{Procedural} & \phantom{0}37 & 23.4\% \\
\bottomrule
\end{tabular}
\caption{Content of the 158 consolidated mined rules over the 15 Market fault
types. Onset placement and candidate arbitration are procedural: they do the
work Step~4 of the five-step procedure does. The per-rule coding is released.}
\label{tab:rule_content}
\end{table}

Seven rules, one in each of seven fault types, countermand the earliest-first criterion
explicitly: ``do not pick the earliest cluster anomaly by default''; ``reject
candidate explanations whose onset is strictly earlier than the target incident
onset''; ``do not use \emph{freshest anomaly} as the onset heuristic''. The
rules contrast and the procedure contrast therefore overlap in content as well
as in repaired cases. The protocol holds the \emph{inputs} fixed, so the arms stay
comparable, but it does not prevent a mined document from encoding procedural
corrections, and that bounds how far the two interventions can be read as
independent.

This also qualifies the procedure result. Removing the five-step prompt leaves
the arm procedure-free only when no rules document is injected; B$'$ with the
mined rules substitutes mined onset and arbitration guidance for the prescribed
kind, so the 42.31 cell is a comparison between two procedures.

\subsection{Human validation of the gap classification}
\label{sec:human_validation}

The gap labels are produced by the same model that made the predictions, so we
checked them against human judgement. Two annotators independently labelled a
stratified sample of the 130 dimensions carrying a gap label, namely all 10 of
the Type~2 dimensions and 20 Type~1 dimensions drawn at random. Both are
practising site reliability engineers at the authors' organisation and are not
authors of this paper. They worked from identical item packs, without access
to each other's responses. Type~2 is
7.7\% of the population and was oversampled to 33\% so that agreement on it
could be measured at all; we report the population-weighted figure alongside the
sample figure for that reason.

Each item showed the annotator the ground truth for one dimension, Stage~1's
prediction, and the evidence the diagnostic had: the extracted anomaly rows and
the log and trace findings. The reconstructed path, the Stage~1 comparison and
the gap verdict were withheld, as were the diagnostic's own summaries of signal
strength. Annotators chose between \emph{the evidence identifies the correct
answer} (Type~1), \emph{it does not} (Type~2), and \emph{unsure}.

\begin{table}[t]
\centering\small
\begin{tabular}{lrrrc}
\toprule
\textbf{Comparison} & \textbf{$n$} & \textbf{Agreement} & \textbf{$\kappa$} & \textbf{95\% CI} \\
\midrule
Annotator A vs annotator B & 27 & 0.852 & 0.625 & $[0.27, 0.91]$ \\
Annotator A vs diagnostic & 28 & 0.893 & 0.704 & $[0.36, 1.00]$ \\
Annotator B vs diagnostic & 28 & 0.964 & 0.920 & $[0.73, 1.00]$ \\
\bottomrule
\end{tabular}
\caption{Human validation of the gap classification on 30 stratified items.
$n$ counts items both sides decided; items marked unsure are excluded from the
pair that contains them. Cohen's $\kappa$ is computed on the enriched
sample, in which Type~2 is 33\% against 7.7\% in the population. Intervals are
percentile bootstrap over items, 20{,}000 resamples; at 30 items they are wide
and overlap.}
\label{tab:human_validation}
\end{table}

All three pairings fall in the same band, $\kappa=0.625$ between the
annotators and $\kappa=0.704$ and $\kappa=0.920$ between the diagnostic and each of
them (Table~\ref{tab:human_validation}). At 30 items the bootstrap intervals are wide
and overlap, so we do not claim the diagnostic agrees with the annotators more
closely than they agree with each other; what the sample supports is that its
labels are not an outlier relative to human judgement on this task. Annotator B confirmed every Type~2 label it
decided, 9 of 9, and annotator A confirmed 5 of 8; weighting by population prevalence the
diagnostic agrees with annotator A on 0.971 of cases and with annotator B on
0.951.

The disagreements concentrate on one fault type. Four of the six items the two
annotators split on are \texttt{container process termination}, where one reads
the joint rise in page faults from zero, small file writes, inode counts and
file descriptors as a recognisable restart and the other does not. Both readings are defensible on the rows shown, so we
read this as a hard case rather than a defect in the classification. It is also
one of the fault types the mined rules do not transfer to (Table~\ref{tab:fault_transfer}).

One limit of the protocol: an annotator cannot issue further tool calls, so the
judgement covers the evidence as shown, whereas the diagnostic could request logs
and traces on demand. The human Type~2 rate is therefore an upper bound on the
rate the diagnostic faces.

\section{Worked Error-Analysis Cases}
\label{sec:worked_cases}

\begin{table}[t]
 \centering
 \small
 \begin{tabular}{lrrr}
 \toprule
 \textbf{Dimension} & \textbf{Correct} & \textbf{Type~1} & \textbf{Type~2} \\
 \midrule
 Component & 33~(47.1\%) & 36~(51.4\%) & \phantom{0}1~(1.4\%) \\
 Reason & 16~(22.9\%) & 46~(65.7\%) & \phantom{0}8~(11.4\%) \\
 Timestamp & 31~(44.3\%) & 38~(54.3\%) & \phantom{0}1~(1.4\%) \\
 \midrule
 Overall$^{*}$ & 14~(20.0\%) & 28~(40.0\%) & \phantom{0}1~(1.4\%) \\
 \bottomrule
 \end{tabular}
 \caption{Gap classification per dimension, Market CB1 rules OFF ($n=70$).
 Type~1 = Reasoning Gap, Type~2 = Data Ambiguity. These are the
 \emph{diagnostic agent's} per-dimension judgements, not the official
 evaluator's scoring points, and the two do not coincide: the diagnostic marks 14 correct overall
because it judges each dimension against the evidence rather than against
exact-match scoring and is not bound by the count-matching rule. \textbf{The evaluator is authoritative for accuracy}.
 $^{*}$Overall additionally includes 27 Mixed cases (38.6\%), those whose gap type differs across dimensions, so the row sums to $n=70$.}
 \label{tab:gap_dim}
\end{table}

Each dimension is classified independently, by checking whether the
discriminating signal was present in Stage~1's retrieved data. A dimension is
\textbf{None} when Stage~1 was already correct on it, and the overall
classification is \textbf{Mixed} when gap types differ across dimensions. The
sources the diagnostic can reach when deciding Type~2 are Stage~1's raw rows, the
inline anomaly tables and the tool results. The Type~1 against Type~2 distinction
rests on one rule: a signal appearing in the (A*) tables exists in the telemetry.
An (A) case is therefore a reasoning failure over data Stage~1 retrieved, and an
(A*) case a failure to retrieve data that was there; neither is a data absence.

Table~\ref{tab:cases} contrasts four cases from the Market CB1 rules OFF
analysis (Section~\ref{sec:error_analysis_results}), one per outcome.
Appendix~\ref{sec:case_dkon} re-runs one of these windows with rules ON.

\begin{table*}[t]
 \centering
 \small
 \setlength{\tabcolsep}{4pt}
 \begin{tabular}{@{}lp{2.6cm}p{2.6cm}p{1.9cm}p{4.6cm}l@{}}
 \toprule
 & \textbf{Ground truth} & \textbf{Stage 1 predicted} & \textbf{GT in candidate pool?}
 & \textbf{Where the discriminating signal was} & \textbf{Gap} \\
 \midrule
 Case 1 & cartservice, network latency
 & node-1, network
 & yes, onset \texttt{1647736740}, exact
 & In Stage~1's own retrieved rows: \texttt{cartservice-grpc mrt} $+1977\%$ at
 \texttt{1647736740}, rising to $+34337\%$
 & Type~1 (A) \\
 \addlinespace
 Case 2 & shippingservice, read I/O load
 & checkoutservice, CPU saturation
 & yes, onset \texttt{1647827640}, exact
 & In the anomaly CSVs (\texttt{container\_fs\_reads} 8338.5 ops, 2410.5\,MB against a
 0.0 threshold); in Stage~1's output only as the phrase ``very large fs reads''
 & Type~1 (A*) \\
 \addlinespace
 Case 3 & cartservice, packet retransmission
 & node-1, disk/IO saturation
 & no
 & Not for this component: 253 extracted rows, none encoding retransmission;
 logs and proxy silent
 & Type~2 \\
 \addlinespace
 Case 4 & productcatalogservice, container CPU load
 & productcatalogservice, CPU throttling
 & yes, selected
 & In Stage~1's own retrieved rows: multi-pod CPU with CFS throttling plus service MRT
 & None \\
 \bottomrule
 \end{tabular}
 \caption{Four representative cases, one per gap outcome. Stage~1
reasons are shown as emitted, before the config agent maps them onto the
allowed vocabulary. In all four the labelled component had extracted anomaly rows; the cases differ in whether Stage~1 surfaced the discriminating
 signal (Case 1, Case 4), surfaced it only as prose (Case 2), or had no such signal in the extracted
 anomalies for that component (Case 3).}
 \label{tab:cases}
\end{table*}

\subsection{Reasoning Gap on unsurfaced data}
\label{sec:case_astar}

Window \texttt{2022\_03\_21/09-30\_10-00}. Stage~1 formed the shippingservice
cluster at the correct onset, emitted it as a candidate, and described its
signature as ``huge CPU usage + throttling, very large fs reads, and
pgfault/pgmajfault bursts''. It then eliminated it in one line, \emph{``later
onset $\rightarrow$ likely downstream amplification''}, and selected
\texttt{checkoutservice}. The numeric rows that would have made the read-I/O
signature explicit were never surfaced (Table~\ref{tab:cases}), although
\texttt{container\_fs\_reads} \emph{was} surfaced numerically in the same
window for \texttt{checkoutservice}, at 431 operations and 15.52\,MB, roughly
$19\times$ and $155\times$ smaller. The failure is therefore not that
read-I/O evidence was unavailable, but that the larger instance was carried
only as prose and so could not be weighed against the smaller one. Rows in
this position are labelled (A*), and the case remains a Reasoning Gap because
the signal existed in the telemetry the agent was given. This is one of the
eight windows from which the \texttt{C1} rule set of
Appendix~\ref{sec:mined_rules} was mined: anchoring the miner on the anomaly
CSVs rather than on Stage~1's output is what keeps that rule recoverable.

\subsection{The other three outcomes}

In Case 1 Stage~1 quantified the decisive \texttt{cartservice-grpc mrt} rows and
emitted the correct candidate at an onset inside tolerance, then eliminated it
as \emph{``mostly symptom (mesh latency, no container resource producer)''} and
selected \texttt{node-1}, whose anomalies were larger in absolute terms and
began earlier. The dismissal rests on cartservice having no independent
producer KPI, but the service-level \texttt{mrt} explosion Stage~1 had just
quantified is exactly such a signal, and node-1's anomalies start before the
labelled cartservice onset, which is inconsistent with treating them as
downstream effects.

Case 3 is the one Type~2 case. cartservice is visibly degraded and the labelled
component has 253 anomaly rows, so extracted evidence is not absent, but
\texttt{system.net.tcp.retrans\_segs} is thresholded cluster-wide and fires in
this window for \texttt{node-1} ($0.30$ against $0.1305$) rather than for
node-5 or node-6, cartservice's hosting nodes. The limitation is threshold
sensitivity on those hosts rather than a property of the telemetry, a
distinction the Type~1/Type~2 split does not capture. This is the only case
among the 70 whose overall classification is Type~2.

Case 4 applies in full the rule the other two misapply: Stage~1 annotated a
candidate pool of six with whether each had an independent producer KPI, and
selected the earliest candidate possessing one rather than the earliest
outright.

\paragraph{The common failure point.} In Case 1, Case 2 and Case 4 Stage~1 generated the
labelled component as an explicit candidate with an onset inside the scoring
tolerance. What separates the correct case from the two incorrect ones is the
producer-KPI clause: in Case 4 it was evaluated and answered affirmatively
(``multi-pod CPU + CFS throttling + service MRT'') and the candidate was kept;
in Case 2 it was not evaluated at all and the candidate was dropped on onset
alone; in Case 1 it was evaluated and answered negatively for a component whose
service-level latency Stage~1 had already measured at $+1977\%$. The failure is
therefore not retrieval, and not the absence of evidence, but a single
elimination step applied inconsistently to evidence already in hand. This is
what the Reasoning Gap category measures, and it suggests making candidate
elimination auditable by requiring the agent to name the producer KPI it
checked and to justify a negative verdict against the rows it has already
enumerated.

\subsection{What the rules change}
\label{sec:case_dkon}

Window \texttt{2022\_03\_21/09-30\_10-00}. This is the case of Appendix~\ref{sec:case_astar}, re-run with rules ON. Stage~1 again formed the shippingservice cluster at \texttt{1647827640}, but this
time recorded its read volume numerically: \texttt{container\_fs\_reads\_MB} of 2410.5, 5642.3
and 3702.5 across consecutive scrapes on \texttt{shippingservice-0}, and 9690.8 rising to
13248.4 on the node-5 replica, against a threshold of 0. It then applied the curated C1
criterion, which requires \texttt{fs\_reads\_MB} above 100~MB per scrape sustained, and rejected
checkoutservice explicitly: its reads are single- or few-timestamp bursts of roughly 4 to 31~MB, ``not consistent
with C1 thresholds''. shippingservice was selected at the correct onset, and the case scores
$1.0$ against $0.0$ in rules OFF.

The rule supplies exactly the comparison Appendix~\ref{sec:case_astar} found missing. In rules OFF
both candidates were present, but only the smaller was quantified and elimination fell back on
onset order alone. What domain knowledge contributes here is not new evidence but a magnitude
criterion that makes the two candidates comparable.

\section{Cases Where Rules Hurt}
\label{sec:dk_hurts}

Injected rules are not monotonically helpful. Table~\ref{tab:transitions} counts, per domain, the
cases that rules OFF answered incorrectly and rules ON answered correctly, and the reverse
(\texttt{results/dk\_transitions.csv}). Manually authored rules break a previously correct answer
in every domain, from 2 cases on Telecom to 10 on Bank; the net gain is the difference between
two sizeable flows rather than a uniform improvement. The mined rules on held-out CB2 have the
most favourable ratio we observe, 19 helped against 2 hurt, and the curated-to-mined transition
on the same cloudbed still costs 3 cases.

\begin{table}[t]
 \centering
 \small
 \begin{tabular}{lrrrr}
 \toprule
 \textbf{Comparison} & \textbf{$n$} & \textbf{Helped} & \textbf{Hurt} & \textbf{Net} \\
 \midrule
 \multicolumn{5}{l}{\textit{Rules OFF $\rightarrow$ manually authored rules}}\\
 \quad Market CB1 & 70 & 10 & \phantom{0}4 & $+6$ \\
 \quad Market CB2 & 78 & 11 & \phantom{0}3 & $+8$ \\
 \quad Bank & 136 & 22 & 10 & $+12$ \\
 \quad Telecom & 51 & 13 & \phantom{0}2 & $+11$ \\
 \quad Market CB1 (Gemma) & 70 & 12 & \phantom{0}1 & $+11$ \\
 \addlinespace
 \multicolumn{5}{l}{\textit{Other knowledge sources, Market CB2}}\\
 \quad Rules OFF $\rightarrow$ mined & 78 & 19 & \phantom{0}2 & $+17$ \\
 \quad Manual $\rightarrow$ mined & 78 & 12 & \phantom{0}3 & $+9$ \\
 \bottomrule
 \end{tabular}
 \caption{Cases whose correctness changes when rules are injected. \textbf{Helped} counts cases
 incorrect without rules and correct with them, \textbf{Hurt} the reverse, over all labelled
 cases $n$ in that domain. Every rule set breaks answers it previously got right.}
 \label{tab:transitions}
\end{table}

\section{Prior-Window Screening Agents}
\label{sec:prior_window_agents}

These agents post-process Stage~1 predictions using prior-window context; all are evaluated only as a GT-oracle ablation (Table~\ref{tab:ablation}) and are not part of the deployed system.

\paragraph{Setup and result.} Table~\ref{tab:ablation} compares the
deployed config-only system, which applies vocabulary enforcement alone,
against an upper-bound variant adding the three agents described below.

\begin{table}[t]
 \centering
 \small
 \resizebox{\columnwidth}{!}{%
 \begin{tabular}{lcccc}
 \toprule
 & \multicolumn{2}{c}{\textbf{Market CB1}} & \multicolumn{2}{c}{\textbf{Market CB2}} \\
 \cmidrule(lr){2-3} \cmidrule(lr){4-5}
 \textbf{Configuration} & \textbf{Full} & \textbf{Partial} & \textbf{Full} & \textbf{Partial} \\
 \midrule
 \multicolumn{5}{l}{\textit{Rules ON}}\\
 \quad config only (system) & 25.71 & 45.60 & 24.36 & 45.63 \\
 \quad + prior-window$^{\dagger}$ & 25.71 & 47.49 & 26.92 & 46.59 \\
 \addlinespace
 \multicolumn{5}{l}{\textit{Rules OFF}}\\
 \quad config only (system) & 17.14 & 35.59 & 14.10 & 28.21 \\
 \quad + prior-window$^{\dagger}$ & 18.57 & 38.57 & 15.38 & 31.09 \\
 \bottomrule
 \end{tabular}}
 \caption{Post-Stage-1 prior-window screening ablation on Market CB1/CB2 (OpenRCA Full / Partial). \emph{config only} is the deployed system. \emph{+ prior-window}$^{\dagger}$ adds the cross-window continuation, recurring-pattern, and revision agents; its cross-window component consults prior-incident ground-truth labels in \textbf{both} modes, so this row is a \textbf{GT-oracle upper bound}, not a label-free system result. Even with oracle information, the tested procedure produces only small numerical Full-Score changes: at most $+2.56$ (CB2, rules ON), with rules OFF changes of $+1.43$ on CB1 and $+1.28$ on CB2. All configurations reuse the same Stage~1 outputs.}
 \label{tab:ablation}
\end{table}

The rules-ON/OFF differences are numerically larger than anything this
screening procedure produces, in either mode;
Section~\ref{sec:statistical_validity} reports which rules comparisons separate from
noise.

\paragraph{Cross-window continuation agent.} Detects whether faults from the preceding 30-minute window carry over into the current window, using a three-condition check:
\begin{enumerate}
 \item \textbf{Prior signal existence}: pre-window raw telemetry shows the component's KPIs elevated before the window boundary.
 \item \textbf{Signal at window start}: anomaly KPIs are elevated at the first one or two timestamps of the current window.
 \item \textbf{Declining trend across the boundary}: pre-window raw values exceed early in-window values.
\end{enumerate}
All three must hold for a prediction to be screened as carry-over. Prior-window anomaly files are generated on demand from the pre-computed thresholds before this check runs. The agent supports a label-free mode (pre-window raw telemetry only) and a GT-oracle mode that additionally consults prior ground-truth window labels; we report only the GT-oracle mode, as an upper bound ($\dagger$ in Table~\ref{tab:ablation}). Such boundary carry-over can arise because failure injection times do not always align with window boundaries.

\paragraph{Recurring pattern detector.} Classifies each KPI in Stage~1's predictions by comparing current-window signals against up to two prior windows as \textsc{Persistent} (same KPI anomalous at same or higher magnitude in prior windows), \textsc{Declining} (anomalous before, now significantly weaker), \textsc{New} (anomalous now but absent in prior windows), or \textsc{Escalating} (anomalous before, now significantly stronger). Predictions whose fault-relevant KPIs are all \textsc{Persistent} are screened out as background.

\paragraph{Revision agent.} Fills prediction vacancies created by either the cross-window or recurring-pattern screening by selecting the next-best candidate from Stage~1's dismissed candidate list, ensuring the prediction count is never reduced below one.

\begin{figure*}[tp]
\centering
\begin{lrbox}{\promptbox}
\begin{minipage}{\dimexpr\textwidth-2\fboxsep\relax}
\begin{minipage}[t]{0.485\textwidth}
\scriptsize
\begin{verbatim}
ALLOWED COMPONENTS: {components} ALLOWED REASONS:
 {reasons}
NUMBER OF ROOT CAUSES TO IDENTIFY: {num_root_causes}
SUPPLEMENTARY TOOLS - call as directed by the METADATA RULES
 above.
 Available: get_mesh_analysis, get_runtime_analysis,
 get_component_traces,
 get_high_latency_traces, get_component_logs,
 get_all_log_errors,
 get_component_proxy_logs, get_all_proxy_errors,
 get_all_trace_errors

ANALYSIS - execute the following steps in order

STEP 1 - ENUMERATE
Read container, node, and service anomalies completely.
These are the primary root cause sources - do not skip or
 summarise any component.
For every anomalous component list: KPI | value | threshold
 | deviation% | timestamp.
Note which node hosts each container (from the cmdb_id
 prefix, e.g. node-6.cartservice-1).

STEP 2 - CLUSTER (uses METADATA for topology)
Form one candidate cluster per distinct logical component:
 - Each service (cartservice, adservice, frontend,...) is
 its own cluster - always.
 Do NOT merge separate services into one cluster because
 they show similar
 symptoms at the same time. Similar simultaneous symptoms
 may indicate a
 cascade, but that is determined in Step 3, not here.
 - For each service cluster: include its container
 anomalies (all pods) +
 the hosting node anomalies + its service-level
 anomalies.
 - Timestamps within a cluster must be within 3 minutes of
 each other.
 A gap of more than 3 minutes = separate clusters
 (separate fault events).
 - Nodes identified as infrastructure in METADATA host no
 services
 - always standalone clusters, never merged with service
 clusters.
List ALL clusters before proceeding. Do not discard any.

STEP 3 - ANALYZE EACH CLUSTER
 [rules ON ] uses RCA RULES + METADATA for tools
 [rules OFF] uses anomaly data + METADATA
For each cluster independently:
 [rules ON ] Apply the RCA RULES to identify fault type
 signals and onset timestamp.
 [rules OFF] Identify the fault type based on the KPI patterns
 in the anomaly data
 (CPU, memory, disk, network, mesh metrics) and
 the onset timestamp.
 Use METADATA tool conditions to decide which supplementary
 tools to call.

 DO NOT decide whether a cluster is a victim or root cause
 here.
 DO NOT dismiss any cluster as a cascade or downstream
 symptom here.
 That determination requires comparing ALL candidates'
 onsets - which happens in Step 4.

 Record for each cluster:
 CANDIDATE: <component> FAULT_TYPE: <type> ONSET_TS:
 <ts>
 CONFIDENCE: high|medium|low LEVEL:
 container|node
 Evidence: <KPIs that matched, with values>
\end{verbatim}
\end{minipage}\hfill
\begin{minipage}[t]{0.485\textwidth}
\scriptsize
\begin{verbatim}
STEP 4 - SELECT
Pool all candidates from Step 3.
 Select the best candidate using the following
 criteria in order:

 1. Temporal ordering (mandatory first):
 List all candidates sorted by onset timestamp, earliest
 first.
 A candidate with an EARLIER onset than all others
 cannot be
 a victim of any of them - it must be kept as a
 candidate.
 Only dismiss candidate A as a victim of candidate B if:
 - A's onset is strictly AFTER B's onset, AND
 - A has no independent producer KPIs of its own.
 Note: mesh latency evidence (grpc duration, grpc error
 codes)
 counts as a producer KPI for network fault
 classification.

 2. Causal independence:
 Do not select two candidates where one fully explains
 the other.

 3. Vocabulary match:
 Component and reason must be from the allowed lists.

Select {num_root_causes} most credible root cause(s).
Justify each selection and each rejection (one line each).

STEP 5 - REFLECTION
 [rules ON ] Review your selection against the RCA RULES
 above: is the cited evidence
 present in the data? Is the fault type correctly
 classified? Are component
 and reason from the allowed vocabulary? If
 anything is inconsistent with
 the rules, revise before writing output.
 [rules OFF] Review your selection for internal consistency:
 is the cited evidence
 actually present in the data? Is the causal
 direction correct
 (cause -> symptom)? Are component and reason from
 the allowed vocabulary?
 If anything is inconsistent, revise before
 writing output.

OUTPUT FORMAT

Candidates found: <list all candidates with evidence>
Final Selection: <candidate pool table> <selection
 reasoning>
 <ruled-out justifications>

Then one prediction block per selected root cause:

=== STAGE 1 PREDICTION ===
Fault Index: <N of {num_root_causes}> Pass:
 PASS_1_UNIFIED
Component: <from allowed list>
Reason: <from allowed list>
Timestamp: <onset timestamp>
Confidence: <high | medium | low> Justification: <1-2
 sentences>
Evidence Summary: <key anomaly rows: component | KPI | value
 | threshold | deviation%>
Fault Type: <fault type from rules>
Duration Pattern: <SPIKE | SUSTAINED | ESCALATING |
 DECLINING>
Instance Scope: <SERVICE-WIDE | POD-LEVEL | NODE-LEVEL>
Causal Chain: <trigger KPI -> downstream effects,
 PRODUCER/SYMPTOM labels>
Secondary Source Findings: <tool findings or "Skipped - not
 needed">
\end{verbatim}
\end{minipage}
\end{minipage}
\end{lrbox}
\colorbox{promptbg}{\usebox{\promptbox}}
\caption{The Stage~1 prompt (Market). Passages that differ between modes are marked
 \textsc{[rules on]} and \textsc{[rules off]}; everything else is identical in both.}
\label{fig:prompt}
\end{figure*}

\begin{figure}[t]
\centering
\begin{lrbox}{\bprimebox}
\begin{minipage}{\dimexpr\columnwidth-2\fboxsep\relax}
\scriptsize
\begin{verbatim}
ANALYSIS

Read the anomaly evidence across all levels before
drawing any conclusion, and identify the root cause.
Call the supplementary tools when the METADATA RULES
above indicate they are needed. Use your own judgement
about how to reason over the evidence; no fixed
procedure is prescribed.
\end{verbatim}
\end{minipage}
\end{lrbox}
\colorbox{promptbg}{\usebox{\bprimebox}}
\caption{The B$'$ analysis block (Market). It replaces \textsc{analysis - execute the
 following steps in order} together with \textsc{step 1} to \textsc{step 5} of
 Figure~\ref{fig:prompt}; B$'$ emits no \textsc{step} headings at all. Everything
 else is identical to rules OFF: the system metadata block, the anomaly data, the
 allowed component and reason lists, the nine-name \textsc{supplementary tools} list,
 and the entire \textsc{output format} block down to its
 \texttt{<fault type from rules>} placeholder, which is a shared template string in both arms.}
\label{fig:prompt_bprime}
\end{figure}

\section{Deployed Pipeline}
\label{sec:pseudocode}

Algorithm~\ref{alg:pipeline} states the deployed system end to end. Phases~1 and~2 run offline and issue no LLM call at query time. The four
detection methods of Phase~1 are \textbf{P95} (values above the 95th percentile,
for KPIs where high values are anomalous), \textbf{P5} (values below the 5th
percentile, where low values are anomalous), \textbf{Boolean} (any non-zero
value where the threshold is zero), and \textbf{ROC} (per-step rate of change
above a learned threshold). A KPI counts as effectively constant, and is
therefore excluded as static, when its variance over the day falls below a small
threshold. The assigning LLM sees each KPI's mean, standard deviation, min/max,
5th/95th percentiles and per-step rate of change, together with descriptions of
the dataset's fault patterns, and infers which direction of deviation is
anomalous. Threshold statistics are computed over the complete telemetry day
containing the evaluated windows, and the resulting per-KPI method--threshold
configurations are stored for reuse. Phase~3 is one Stage~1 call per window, conditional sub-agent
calls, and one config-agent call; the prior-window screening agents of
Appendix~\ref{sec:prior_window_agents} are evaluated only as an ablation and are not part of this
path. \textsc{Rules} is empty under rules OFF, the manually authored document under rules ON, and the
mined document in the mined configuration. Every other input is identical across the three, except
that three prompt passages in Steps~3 and~5 are conditioned on whether a rules
document is present, as marked in Figure~\ref{fig:prompt}; Step~4's selection
criteria are identical in both modes.

\begin{algorithm}[t]
\small
\caption{Deployed reasoning pipeline}
\label{alg:pipeline}
\begin{algorithmic}[1]
\Statex \textbf{Offline, once per dataset and day}
\For{each metric source $s$ and KPI $k$ in $s$}
 \If{$k$ is effectively constant over the day}
 \State mark $k$ \textbf{static}; exclude
 \Else
 \State $m_k \gets$ \Call{AssignMethod}{summary statistics of $k$}
 \Comment{P95 / P5 / Boolean / ROC; one LLM call per source-day}
 \State $\theta_k \gets$ threshold for $m_k$ from that day's distribution
 \EndIf
\EndFor
\Statex
\Statex \textbf{Per 30-minute query window $W$}
\State $A \gets \{$rows of $W$ where $v_k$ crosses $\theta_k\}$ for every dynamic $k$
 \Comment{scripted; no LLM call}
\State $L, T \gets$ full log and trace slices of $W$ \Comment{not thresholded}
\Statex
\State $P \gets \textsc{Metadata} \cup \textsc{Vocabularies} \cup \{\,|R|\,\}
 \cup \textsc{ToolSignatures} \cup \textsc{Rules}$
\State $y \gets \textsc{Stage1}(A, P, L, T)$ \Comment{one reasoning call:
 enumerate, cluster, analyse, select, reflect. It invokes the log, trace and
 (Market) mesh and runtime tools as the metadata gating conditions direct;
 their output returns to this call. Log and trace retrievals are scripted,
 the mesh and runtime sub-agents are one LLM call each}
\State $\hat{y} \gets \textsc{ConfigAgent}(y, \textsc{Vocabularies})$
 \Comment{one call: map free text onto allowed labels}
\State \Return $\hat{y}$ as $\langle$component, reason, onset$\rangle \times |R|$
\Statex
\Statex \textbf{Scoring}
\State score $\hat{y}$ with OpenRCA's evaluator; a case scores $0$ if
 $|\hat{y}| \neq |R|$
\end{algorithmic}
\end{algorithm}

\section{Stage 1 Prompts}
\label{sec:prompt_stage1}

Figure~\ref{fig:prompt} gives the Stage~1 prompt for Market. The system metadata block
(Appendix~\ref{sec:dk_artifacts}) is injected above the analysis steps in both modes, and the
curated rules block only in rules ON. The Telecom and Bank prompts follow the same five-step structure and
differ only in their tool lists (Appendix~\ref{sec:tools}) and fault
vocabularies. Figure~\ref{fig:prompt_bprime} gives the B$'$ analysis
block, which is the only part of the prompt that differs in that arm.

\section{Domain Knowledge Artifacts}
\label{sec:dk_artifacts}

Three artifacts are injected into the Stage~1 prompt. System metadata and the component and reason vocabulary are injected in both modes, so
the rules flag does not control them. They are not, however, free of diagnostic content, and we
describe the metadata in full here rather than by size alone.

Telecom's metadata opens with a topology section listing component
names, tier ordering and upstream--downstream relationships. The routing
rules state which metric files are primary rather than secondary
evidence, and the gating conditions which log and trace tools may be
invoked per component type. The Market metadata document has four parts. First, system facts: which hosts
are infrastructure nodes and therefore run no services. Second, data-source routing keyed to fault
type, for example that mesh anomalies are relevant for network-related issues and are not to be
used as a primary signal for CPU, memory or disk faults, and that the runtime source records only
one service. Third, a nine-row table giving the condition under which each supplementary tool
should be called, again keyed to fault type. Fourth, component-name granularity rules: strip the
\texttt{<node>.} prefix from \texttt{cmdb\_id}, report a single affected pod by its own name, and
report the service name when three or more of a service's four pods are affected simultaneously.

The second and fourth parts are discrimination knowledge, and the fourth bears directly on
scoring, since OpenRCA matches component names exactly. What the rules flag isolates is the \emph{fault-discrimination rule
document}, with the metadata held constant on both sides, so the rules-OFF against rules-ON
contrast is unaffected by this; what changes is the description of the baseline, which is a
reduced-knowledge condition rather than a knowledge-free one. Two consequences are reported where
they arise: the granularity rule is an input our pipeline has and RCLAgent does not
(Appendix~\ref{sec:llm_failure_analysis}), and because the tool-invocation table lives in the
metadata, the tools and the metadata cannot be ablated independently in this architecture. The two documents rest on different amounts of supervision,
25 of 57 CB1 windows against reports from all 70 cases, so supervision
volume is confounded with provenance.

The curated rule documents are injected only in rules ON, and in the mined-rules configuration the Market document is replaced by the automatically mined one. The full text of every artifact is included in the supplementary material under \texttt{domain\_knowledge/}.

The three documents are organised differently, reflecting how each was written. The Market rules
are the most systematic: ten numbered parts covering a fault-type taxonomy (15 types: 9 container
and 6 node), source selection per fault type, container resource/process classification
(C1--C5), container network classification (C6--C9), node classification (N1--N6), cross-layer rules,
special handling for process termination, ambiguity handling, scan order, and classification
principles. The Telecom document is instead written as a set of dataset-specific overrides on a
general rule base, with each override marked. Both state their provenance
explicitly: the Telecom rules describe themselves as reverse-engineered from the
ground-truth labels of roughly 24 analysed cases, of which the recoverable
case-study logs identify 20 (Table~\ref{tab:heldout}), and the Bank rules as empirically derived from ground-truth
labels in that dataset, the developed-on-test exposure quantified in
Table~\ref{tab:heldout}.

\section{Investigation Tools and Sub-Agents}
\label{sec:tools}

Table~\ref{tab:tools} lists the investigation tools available in each domain; the sets differ,
reflecting which telemetry modalities each domain provides.
Tools are gated: the system metadata specifies a calling condition per tool, and agents are instructed not to call them pre-emptively. The sets are fixed rather than written by the model while it reasons. Market additionally exposes two LLM sub-agents that
analyse the mesh and runtime metric sources on demand; Telecom and Bank have no sub-agents.

\begin{table}[t]
 \centering
 \small
 \begin{tabular}{lp{4.6cm}}
 \toprule
 \textbf{Domain} & \textbf{Tools} \\
 \midrule
 Market (9)
 & \emph{sub-agents:} \texttt{get\_mesh\_analysis}, \texttt{get\_runtime\_analysis};
 \emph{logs:} \texttt{get\_component\_logs}, \texttt{get\_all\_log\_errors},
 \texttt{get\_component\_proxy\_logs}, \texttt{get\_all\_proxy\_errors};
 \emph{traces:} \texttt{get\_component\_traces},
 \texttt{get\_high\_latency\_traces}, \texttt{get\_all\_trace\_errors} \\
 \addlinespace
 Telecom (7)
 & \texttt{get\_container\_traces}, \texttt{get\_error\_spans}, \texttt{get\_db\_spans},
 \texttt{get\_node\_spans}, \texttt{get\_volume\_by\_minute},
 \texttt{get\_bimodal\_check}, \texttt{get\_all\_spans\_summary} \\
 \addlinespace
 Bank (8)
 & \emph{logs:} \texttt{get\_gc\_logs}, \texttt{get\_all\_gc\_events},
 \texttt{get\_access\_logs};
 \emph{traces:} \texttt{get\_component\_spans}, \texttt{get\_high\_latency\_spans},
 \texttt{get\_span\_count\_by\_component},
 \texttt{get\_intermittent\_spike\_pattern}, \texttt{get\_caller\_spans} \\
 \bottomrule
 \end{tabular}
 \caption{Investigation tools per domain. Telecom is trace-only because the domain provides no
 log data. Bank exposes JVM garbage-collection and access-log tools not present elsewhere.}
 \label{tab:tools}
\end{table}

\section{Reproducibility}
\label{sec:reproducibility}

All primary results in Table~\ref{tab:system} use the \texttt{gpt-5.2} model identifier with
\texttt{max\_completion\_tokens} $=32000$ and \texttt{temperature} $=0$; the explicitly labelled
Gemma secondary-model results are reported separately in Table~\ref{tab:gemma}. Every agent in the
pipeline, namely Stage~1, the mesh and runtime sub-agents, the config agent and the prior-window
screening agents evaluated in the ablation, uses the same setting; no sampling is used anywhere.
Greedy decoding is not reproducible across serving, however, so repeated
runs of the same arm on the same inputs need not be identical.

The deployed system issues two sequential LLM calls per
window: one Stage~1 reasoning call and one config-agent call. The log- and
trace-investigation tools Stage~1 calls on demand are scripted retrievals that
add turns to that call rather than further LLM calls. Market additionally issues
a mesh or runtime sub-agent call, or both, when the fault type triggers the
corresponding gating conditions, so the per-window count is two on Telecom and
Bank and two to four on Market. Anomaly pre-computation (Phases~1 and~2) is scripted and requires no LLM calls at query
time, and the error-analysis and rule-mining pipelines are
diagnostic and add no production cost.

\paragraph{Code and artifacts.} The supplementary material provided for review contains the offline anomaly pipeline of Phases~1 and~2
(\texttt{anomaly\_pipeline/}); the Phase~3 reasoning pipeline as one package per domain
(\texttt{forward\_reasoning\_\{market,telecom,bank\}\_final/}); the artifacts injected into the
Stage~1 prompt, namely system metadata, curated rules and the allowed vocabularies
(\texttt{domain\_knowledge/}); the reverse-reasoning and gap-analysis code
(\texttt{error\_analysis/}); the rule-mining pipeline together with the mined rule sets
(\texttt{rule\_mining/}); the label-and-schema control document, its fifteen per-fault-type
generations and the per-window outputs of its end-to-end CB2 evaluation
(\texttt{control\_rules/}); the scoring scripts that produce the reported Full and Partial scores
and the reduced-subset sensitivity re-evaluation (\texttt{evaluation/}); the per-window Stage~1 outputs for every reported configuration, including the secondary-model, mined-rules and prior-window ablation runs (\texttt{results/forward\_reasoning/}); the reverse-reasoning reports behind the error analysis (\texttt{results/reverse\_reasoning/}); the baseline-analysis code
(\texttt{baselines/}), including the diagnostic GALA reconstruction and our corrected
preprocessing, execution and evaluation wrappers around the causal-discovery implementations
provided by RCAEval; and the B$'$ ablation harness together with the official-evaluator scoring
wrapper and the paired significance tests (\texttt{ablation\_bprime/}). The OpenRCA dataset itself is not redistributed.

\section{Supporting Measurements}
\label{sec:supporting_measurements}

Table~\ref{tab:gemma} reports the secondary-model evaluation discussed in Section~\ref{sec:ablation}, and Table~\ref{tab:heldout} the leakage-control re-evaluation discussed in Section~\ref{sec:leakage_control}.

\noindent\textbf{Secondary model.} Gemma-4-31b-it scores below GPT-5.2 on
Market CB1 without rules, as does the OpenRCA agent on the same pair; with
rules injected the two models tie on Full Score and Gemma trails on Partial.
The gain from rules is larger for the weaker model, $+15.71$ Full against
$+8.57$, consistent with explicit rules supplying something a stronger model
already supplies implicitly, though both arms are single runs whose gains are 11
and 6 cases, and we do not test that reading.

\noindent\textbf{Solved-case counts.} Table~\ref{tab:system} reports percentages over the
case counts of Table~\ref{tab:dataset_counts}, namely 70 on Market CB1, 78 on CB2,
136 on Bank and 51 on Telecom. In the same column order the counts are 8, 13,
38 and 15 for the OpenRCA agent; 12, 11, 38 and 18 for rules OFF; and 18, 19,
50 and 29 for rules ON. On the two cloudbeds where it is run, B$'$ solves 19 and 22
without rules. On held-out CB2 the mined rules solve 28, B$'$ with the mined
rules 33, and the label-and-schema control 15. The onset-ordering arm of
Section~\ref{sec:ablation} solves 19 on CB2.

\begin{table}[t]
 \centering
 \small
 \begin{tabular}{lcc}
 \toprule
 \textbf{Configuration} & \textbf{Full} & \textbf{Partial} \\
 \midrule
 OpenRCA agent & 4.29 & 17.24 \\
 Ours, rules OFF & 10.00 & 26.54 \\
 Ours, rules ON & 25.71 & 43.21 \\
 \bottomrule
 \end{tabular}
 \caption{Gemma-4-31b-it model on Market cloudbed-1 (OpenRCA Full / Partial).}
 \label{tab:gemma}
\end{table}

\begin{table}[h]
 \centering
 \small
 \setlength{\tabcolsep}{3pt}
 \begin{tabular}{lcccccc}
 \toprule
 \textbf{Domain} & \textbf{ON full} & \textbf{ON h/o} & \textbf{$\Delta$}
 & \textbf{OFF h/o} & \textbf{Matched $\Delta$} & \textbf{$p$} \\
 \midrule
 Telecom & 56.86 & 51.61 & $-5.3$ & 22.58 & $+29.03$ &.0117 \\
 Market CB1 & 25.71 & 23.53 & $-2.2$ & 17.65 & $+5.88$ &.7266 \\
 Bank & 36.76 & 38.14 & $+1.4$ & 28.87 & $+9.28$ &.1078 \\
 \bottomrule
 \end{tabular}
 \caption{Matched held-out evaluation of the manually authored rules. Both arms are rescored on
 the same reduced subset, which excludes the windows used to construct that domain's rules, so
 \textbf{Matched $\Delta$} is a like-for-like rules ON/OFF effect rather than a sensitivity check.
 Held-out case counts are 31 of 51 for Telecom, 34 of 70 for Market CB1 and 97 of 136 for Bank.
 $p$ is exact McNemar on the paired cases; these three form a separate confirmatory family and
 under Holm within it Telecom survives at $.0351$ while the other two do not. On the excluded
 construction windows the same contrast is $+10.00$ points for Telecom (OFF 11/20, ON 13/20),
 $+11.11$ for Market CB1 (6/36, 10/36) and $+7.69$ for Bank (10/39, 13/39), so Telecom's larger
 held-out gain reflects rules OFF degrading on the harder subset rather than rules ON holding up,
 and only Market shows the split a leakage effect would produce. \textbf{ON h/o} against \textbf{ON full} is readable as a one-arm sensitivity check. The exclusion sets differ in kind: the Telecom and Bank sets are recovered
 from case-study logs that name each analysed window with its ground truth, removing 20 of 51 Telecom cases and 39 of 136 Bank cases, whereas the Market rules carry no per-window citations:
the construction count is known but the individual windows are not, so the Market set is a proxy, the first two
windows of each of the 15 fault reasons, removing 36 of the 70 cases; it is
larger than the construction set but not guaranteed to contain it. Market CB2 is omitted because no CB2
 windows were used to construct the Market rules. Reported numbers are OpenRCA Full Scores. Model
 identifier: GPT-5.2.}
 \label{tab:heldout}
\end{table}

\noindent\textbf{What the Telecom held-out margin rests on.} The
Telecom margin rests on 11 discordant cases split 10 to 1, so one case moving the other way raises the exact
value to $.0654$ and the Holm value within this family to $.196$. The construction set is also only partly recoverable: the
Telecom rules cite roughly 24 analysed cases and the logs identify 20, so up to
4 may remain among the held-out 31. If all four are cases only rules ON
solves, the matched margin falls from $+29.03$ to $+18.52$ and the exact value
rises to $.125$. The margin is therefore robust in size and fragile in
significance, and we read the Telecom held-out result as directional.

\noindent\textbf{Threshold sweep.} Sweeping the extraction threshold from P90 to P99 changes
anomaly-row volume by an order of magnitude. At the deployed P95 setting, near-onset component
coverage is 92.9\% and coverage of KPIs cited as discriminating by the mined rules is 91.4\%. The labelled component has metric anomaly rows at all in 69 of the 70 cases (98.6\%), unchanged across every setting swept. These figures cover the thresholded metric sources, exclude logs and traces, and apply both value tails to every KPI where the deployed detector assigns one method per KPI, so they are upper bounds. Across the sweep these move from 94.3\% to 84.3\% and from 92.9\% to 80.0\% respectively. This
measures retention of signals the rules rely on, not Full Score at each threshold, which we did
not evaluate.

Every score is a single run except the CB2 rules-OFF arm,
which was re-run once and scored 12 of 78 against 11 of 78,
with one case of the 78 changing between the two runs. Serving non-determinism
under greedy decoding remains largely unmeasured and could move individual
contrasts by a few cases.

\noindent\textbf{Tokens and cost.} Token usage was not captured for the GPT-5.2 \texttt{main}
arms, which predate the accounting module. It is captured for the Gemma path and for B$'$, where
a representative CB1 window logged 151{,}849 input and 22{,}280 output tokens over two calls. The prices embedded in that module are the Gemma rates, so any cost it printed during a GPT-5.2 run is not a valid figure. The mined-rules configuration is one of the \texttt{main} arms, so the additional input tokens its rule document contributes were not measured either.

\section{Statistical Analysis}
\label{sec:statistical_analysis}

\noindent\textbf{Onset-anchor pattern.} Under rules OFF, 32 of 34 erroneous
onsets are early, the median absolute onset error is 468\,s, and 44\% of predicted
onsets fall on a 30-minute window boundary where no injected onset can lie, against
24\% and 49\,s for B$'$; rules ON shows neither pattern (53\% onset accuracy,
24\% boundary rate, 44\,s median error). On held-out CB2 with the mined
rules the boundary rate is 20 of 108 predicted onsets
(18.5\%) with the procedure and 16 of 108 (14.8\%) without it, the
denominator counting every predicted onset rather than one per case, so the pattern is
absent whether or not the procedure is present once rules are. These are descriptive rates and we attach no test: at those
counts the two CB2 arms are not separable. Removing only the onset-ordering
criterion, with the rest of the procedure intact, moves CB2's onset behaviour to
where B$'$ sits on CB1: the boundary rate falls from 53\% to 15\% and the
median absolute onset error from 700\,s to 34\,s, against 24\% and 49\,s for
B$'$ on CB1. The paper reports no CB2 onset figures for B$'$ itself.
Against rules OFF the arm helps 11 cases and hurts 3, an exact McNemar value
of $.0574$; it is not among the declared families and we attach no corrected
verdict. This is an intervention on the
criterion rather than a comparison across arms, so for CB2 the boundary
pattern is no longer only an association. On CB1, moving from rules OFF to B$'$, component accuracy
moves from 52\% to 57\% and onset from 38\% to 53\%, while fault-reason
accuracy falls from 35\% to 29\%. These are CB1 figures and do not describe the
single-criterion arm.

\noindent\textbf{The B$'$ ablation.} Over the 70 labelled cases of Market CB1,
spanning all 57 windows, B$'$ solves $19/70$ against rules OFF's $12/70$, with 13 discordant cases, 10
of which only B$'$ solves. The window-level permutation over the 57 windows, 13
of which hold two cases, gives $p=.149$ against a case-level exact McNemar $.092$; the dependence-aware test is the more conservative of the two.

\noindent\textbf{The CB2 procedure comparison.} On held-out CB2 with the
mined rules in place, B$'$ solves $33/78$ against the mined-rules arm's $28/78$, with 15 discordant cases, 10 of which only B$'$ solves;
the window-level permutation over the 63 CB2 windows gives $p=.412$
against a case-level exact McNemar $.302$, and the window-clustered 95\% CI of $[-5.26, +18.42]$ spans zero.

\noindent\textbf{The remaining CB2 cells.} On held-out CB2 without rules,
B$'$ solves $22/78$ against rules OFF's $11/78$, with 17 discordant cases over
15 windows, 14 of which only B$'$ solves; the window-level permutation gives
$p=.0248$ against a case-level exact McNemar $.0127$, with a window-clustered
95\% CI of $[+3.75, +25.30]$. Adding the mined rules to B$'$ moves it from
$22/78$ to $33/78$, with 17 discordant cases over 12 windows, 14 of which
only the mined arm solves; the permutation gives an exact $p=.0508$
against a case-level exact McNemar $.0127$, with a CI of $[+1.37, +26.32]$ from
20{,}000 resamples. The permutation is enumerated over all $2^{12}$ sign
patterns of the 12 windows with a nonzero difference, so no resampling seed
enters it; the value sits just above the conventional threshold and we attach
no verdict to which side of it falls.

\noindent\textbf{Field-level comparisons.} For the B$'$ analysis we
compare component, reason and onset accuracy across three arm pairs, giving
nine comparisons: rules OFF against B$'$ tests the fixed procedure, since the
two arms differ in the five-step block; rules OFF against rules ON tests the fault-discrimination rules on a shared procedure; and rules ON against B$'$
differs in both and is included only as a descriptive contrast. None of the
nine survives Holm correction. Under the governing test the nine range from $p=.078$ to $p=1.0$. The
procedure comparison on onset, rules OFF against B$'$, gives $p=.173$, with a
window-clustered 95\% CI of $[-.31,+.02]$ that spans zero; its case-level exact
value is $.057$, so the dependence-aware test moves it further from significance
rather than closer.

\noindent\textbf{Field-level denominators and intervals.} Three points on how these are computed. The denominators are 54 or 55 rather than 70 because they count \emph{scoring points} of each kind rather than cases: cases define different scoring fields, and a case whose window produced no prediction still contributes its points as failures. The intervals come from a bootstrap that resamples the 57 evaluation windows rather than the scoring points, so cases sharing a window are not treated as independent. Each interval is on the paired difference in the stated order, first arm minus second.

Table~\ref{tab:discordance} gives the discordance structure behind each
row, so every $p$-value can be checked against the number of windows the
discordant cases occupy rather than the number of cases alone.

\begin{table}[t]
\centering\footnotesize
\resizebox{\columnwidth}{!}{%
\begin{tabular}{@{}lrrrrr@{}}
\toprule
\textbf{Comparison} & \textbf{Cases} & \textbf{Windows} & \textbf{Disc.} & \textbf{Disc.\ win.} & \textbf{Case-level $p$} \\
\midrule
CB2 rules OFF $\rightarrow$ mined & 78 & 63 & 21 & 18 &.0002 \\
Telecom rules OFF $\rightarrow$ manual & 51 & 51 & 15 & 15 &.0074 \\
CB2 control $\rightarrow$ mined & 78 & 63 & 19 & 16 &.0044 \\
Bank rules OFF $\rightarrow$ manual & 136 & 127 & 32 & 31 &.0501 \\
CB2 rules OFF $\rightarrow$ manual & 78 & 63 & 14 & 13 &.0574 \\
CB2 manual $\rightarrow$ mined & 78 & 63 & 15 & 12 &.0352 \\
CB1 rules OFF $\rightarrow$ manual & 70 & 57 & 14 & 13 &.1796 \\
\bottomrule
\end{tabular}%
}
\caption{Discordance structure behind Table~\ref{tab:paired}. \textbf{Disc.}
counts cases where exactly one arm scores Full; \textbf{Disc.\ win.} is the
number of distinct evaluation windows those cases occupy. Where the two differ,
discordances share a window and contribute one sign flip rather than two, which
is why the window-level permutation $p$ departs from an exact McNemar test on
the same discordance count. Telecom has one labelled case per window, so the two tests coincide there.
\textbf{Case-level $p$} is the exact two-sided McNemar value implied by the
discordance split, so every row of Table~\ref{tab:paired} can be checked in
closed form against its governing permutation value. The per-arm splits
reproduce the helped and hurt counts of Table~\ref{tab:transitions}.}
\label{tab:discordance}
\end{table}

\noindent\textbf{Partial Score as a secondary outcome.} Full Score is the
primary outcome throughout and the one every comparison above is tested on. It
is binary per case, so it discards the partial credit the evaluator awards when
some requested fields are correct. After observing that two of the three CB2
rules comparisons do not separate on Full Score, we recomputed all seven rules comparisons on Partial
Score, the mean per-case score, under the same window-level permutation and
the same clustered resampling, correcting within that set. The order matters and we state it: this is a
secondary analysis chosen after seeing the primary result, not a pre-specified
one. Against rules OFF the mined rules gain $+26.82$ points, with a 95\% interval of $[+17.19, +36.59]$ and an
uncorrected $p<.0001$; against the label-and-schema control they gain
$+23.29$, interval $[+13.42, +33.33]$, uncorrected $p<.0001$. Against the manually authored rules they gain $+9.40$, with an interval of
$[+0.54, +18.36]$ and an exact permutation value of $.0498$, enumerated over
all $2^{24}$ sign patterns of the 24 windows with a nonzero Partial
difference. It sits within $.0002$ of the conventional threshold, so we
report it without a significance verdict.

\noindent\textbf{Test and interval share one dependence model.} Because 13
of Market CB1's 57 windows carry two cases, every test is a window-level
sign-flip permutation and every interval a bootstrap over windows. They share
that model but not the procedure, so an interval and its $p$ can disagree at
these window counts; we do not read an interval as a test. Exact McNemar is a
sensitivity check, and the more conservative of the two governs each claim.
The procedure ablation of Section~\ref{sec:ablation} has a 95\% CI of
$[0.00, +21.33]$ on CB1 with $p=.149$; the three CB2 intervals are given above.
Two of the four exclude zero, CB2 without rules and the mined-rules addition to
B$'$; the CB1 interval has zero as its lower bound and the CB2 with-mined-rules
interval spans it.

\noindent\textbf{Families and testing policy.}
Section~\ref{sec:statistical_validity} states the governing test; its
mechanics are a sign-flip permutation over the evaluation windows, with cases
sharing a window flipped together. We declare four families and correct within each: the seven
Full-Score rules comparisons of Table~\ref{tab:paired}, the same seven on Partial Score as
a secondary outcome, the nine field-level B$'$ comparisons above, and the
three matched held-out comparisons of Table~\ref{tab:heldout} declared in
Section~\ref{sec:leakage_control}. The held-out family is kept apart because
every member re-tests a comparison already declared among the seven, on a
subset of that comparison's own cases, against the single objection that the
rules were written against the windows they were scored on. Pooling it with
the seven would give ten tests, and Holm over those ten leaves one comparison
significant, mined versus rules OFF on CB2 at $.0110$, and nothing else. Its
reported $p$ is exact McNemar on the paired cases, which no verdict turns on,
since Telecom is the only survivor there while Bank ($.1078$) and Market CB1
($.7266$) are far from any Holm threshold. The mined-against-control contrast
is the only test-choice disagreement among the seven, at Holm $.0264$
case-level against $.0605$ window-level, and there is none among the nine
(Table~\ref{tab:discordance}); where the two tests differ numerically the
largest movement toward significance is from $.1796$ to $.1431$. Intervals
come from the same clustered resampling rather than Wilson intervals, since a
Wilson interval assumes independent Bernoulli trials that the two-case CB1
windows violate. We report no per-case dispersion table: for a binary
per-case score the standard deviation is fixed by the accuracy at $\sqrt{p(1-p)}$. The complete
nine-row field-level output is included with the B$'$ analysis artifacts under
\texttt{ablation\_bprime/}.

\end{document}